\documentclass{article}

\usepackage{microtype}
\usepackage{graphicx}
\usepackage{subfigure}
\usepackage{booktabs} 

\usepackage{makecell}
\usepackage[dvipsnames]{xcolor}

\usepackage{hyperref}

\usepackage[accepted]{icml2025}

\usepackage{amsmath}
\usepackage{amssymb}
\usepackage{mathtools}
\usepackage{amsthm}

\usepackage[capitalize,noabbrev]{cleveref}

\theoremstyle{plain}

\theoremstyle{definition}

\theoremstyle{remark}

\usepackage[textsize=tiny]{todonotes}

\icmltitlerunning{Complex Wavelet Mutual Information Loss: A Multi-Scale Loss Function for Semantic Segmentation}

\begin{document}

\twocolumn[
\icmltitle{Complex Wavelet Mutual Information Loss: A Multi-Scale Loss Function for Semantic Segmentation}




\begin{icmlauthorlist}
\icmlauthor{Renhao Lu}{cornelBME}
\end{icmlauthorlist}

\icmlaffiliation{cornelBME}{Meinig School of Biomedical Engineering, Cornell University, Ithaca, NY, USA}

\icmlcorrespondingauthor{Renhao Lu}{rl839@cornell.edu}

\icmlkeywords{Semantic segmentation, wavelet transform, steerable pyramid, mutual information}

\vskip .3in
]



\printAffiliationsAndNotice{}  

\begin{abstract}
Recent advancements in deep neural networks have significantly enhanced the performance of semantic segmentation. However, class imbalance and instance imbalance remain persistent challenges, where smaller instances and thin boundaries are often overshadowed by larger structures. To address the multiscale nature of segmented objects, various models have incorporated mechanisms such as spatial attention and feature pyramid networks. Despite these advancements, most loss functions are still primarily pixel-wise, while regional and boundary-focused loss functions often incur high computational costs or are restricted to small-scale regions. To address this limitation, we propose the complex wavelet mutual information (CWMI) loss, a novel loss function that leverages mutual information from subband images decomposed by a complex steerable pyramid. The complex steerable pyramid captures features across multiple orientations and preserves structural similarity across scales. Meanwhile, mutual information is well-suited to capturing high-dimensional directional features and offers greater noise robustness. Extensive experiments on diverse segmentation datasets demonstrate that CWMI loss achieves significant improvements in both pixel-wise accuracy and topological metrics compared to state-of-the-art methods, while introducing minimal computational overhead. Our code is available at \href{https://github.com/lurenhaothu/CWMI}{https://github.com/lurenhaothu/CWMI}
\end{abstract}

\section{Introduction}
\label{Introduction}

Semantic segmentation, the process of partitioning an image into regions associated with semantic labels, plays a crucial role in applications ranging from autonomous driving to biomedical imaging. Despite significant progress driven by deep neural networks such as U-Net \cite{ronneberger2015u} and fully convolutional networks \cite{long2015fully}, challenges persist. Class imbalance, where dominant classes overshadow smaller ones, and instance imbalance, where small-scale structures are frequently ignored, remain major obstacles \cite{jiang2024multi,kofler2023blob}. Addressing these imbalances requires not only pixel-wise accuracy, but also the preservation of structural similarity, a property vital for ensuring spatial coherence and topological integrity in segmented outputs \cite{wang2004image}.

Although the advancements in feature extraction, such as feature pyramid networks \cite{lin2017feature} and attention mechanisms \cite{woo2018cbam,chen2021transunet,islam2020brain}, have enabled models to capture multiscale contextual information, most loss functions still focus on pixel-wise optimization \cite{azad2023loss}. Region-based loss functions, such as Dice loss \cite{milletari2016v} and Tversky loss \cite{salehi2017tversky}, are not subject to class imbalance, but smaller instances within the same class are still easy to be overshadowed, which can be critical for preserving regional and boundary details. Zhao et al. developed Regional Mutual Information (RMI) loss, which captures statistical relationships over regions. However, RMI is constrained within a relatively small region ($3\times3$ pixels) to avoid high computational overhead \cite{zhao2019region}. Thus, a loss function that balances computational efficiency with the ability to model structural and regional dependencies at larger scales is highly desirable.

Addressing these challenges from a frequency-domain perspective provides an innovative pathway. Patterns of varying scales in images are inherently tied to their frequency components: large-scale structures correspond to low frequencies, while finer details correspond to high frequencies. Wavelet transforms are uniquely suited for this multiscale decomposition, as they preserve both spatial and frequency information \cite{mallat1989theory}. Among these, the steerable pyramid, proposed by \cite{simoncelli1992shiftable}, leverages steerable filters for redundant wavelet decomposition, enabling multiscale and multi-orientation feature extraction. Its extension, the complex steerable pyramid, further enhances this framework by using complex numbers to explicitly represent local phase information, allowing for robust extraction of structural details across scales and orientations \cite{portilla2000parametric}. These properties make it a powerful tool for segmentation tasks where structural similarity is paramount.

In this paper, we introduce Complex Wavelet Mutual Information (CWMI) loss, a novel loss function that leverages the complex steerable pyramid for efficient multiscale structural information extraction. By combining the robust multiscale decomposition capabilities of the complex steerable pyramid with the statistical power of mutual information, CWMI loss explicitly incorporates local phase, orientation, and structural features into the loss calculation. This approach ensures structural coherence and boundary preservation while maintaining computational efficiency, making it well-suited for segmentation tasks with significant class and instance imbalances.

Our contributions are summarized as follows:
\begin{itemize}
    \item We propose CWMI loss, which can maximize the mutual information in the domain of complex steerable pyramid decompositions. Such a strategy can enhance multiscale structural features for semantic segmentation, especially for tasks with significant class and instance imbalances.
    \item We demonstrated the superiority of CWMI with extensive experiments on four public segmentation datasets: SNEMI3D (neurite segmentation in electron microscopy slices), GlaS (gland segmentation in H\&E slices), DRIVE (retinal vessel segmentation in fundus images), and MASS ROAD (road segmentation from aerial imagery). Compared with 11 state-of-art (SOTA) loss functions, CWMI showed better performance on both pixel-wise metrics and topological metrics, while introducing minimal computational overhead. 
\end{itemize}

\section{Related Work}

Semantic segmentation has seen tremendous advancements through deep learning architectures, with U-Net and its variants becoming a cornerstone of this field. The U-Net model, proposed by \cite{ronneberger2015u}, utilizes a symmetric encoder-decoder architecture with skip connections to preserve spatial details while capturing global context. Its success has inspired numerous adaptations in its convolutional blocks \cite{huang2019fixed,diakogiannis2020resunet} and skip connections \cite{zhou2018unet++,chen2021transunet}. Inspired by the transformer model, the attention mechanism has also been incorporated into the U-Net structure, which has shown significant performance enhancement, as in Attention U-Net \cite{islam2020brain}, TransUNet \cite{chen2021transunet}, Vision Mamba UNet \cite{ruan2024vm}, etc. In this study, we compare our proposed CWMI loss using U-Net and Attention U-Net to test the generalization and superiority of CWMI, while the general idea is adaptable to other architectures as well. 

\begin{figure*}[!h]
    \centering
    \includegraphics[width=.8\linewidth]{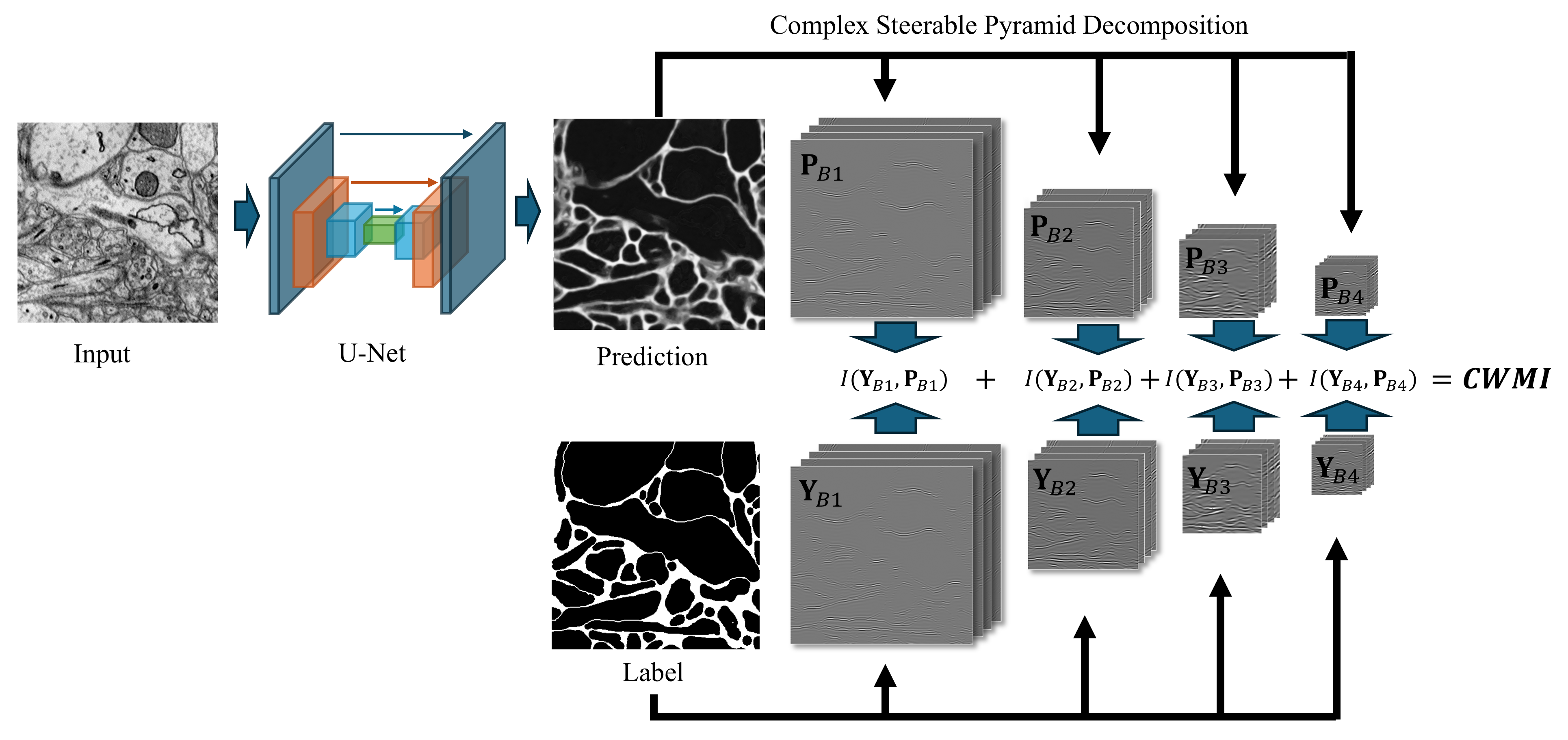}
    \vspace{-10pt}
    \caption{Illustration of the proposed Complex Wavelet Mutual Information (CWMI) Loss. The prediction and label images are decomposed using a complex steerable pyramid, which generates subbands at different scales and orientations. Mutual information is calculated for each corresponding pair of subbands, and the CWMI is computed as the sum of these mutual information values. \(\mathbf{Y}_{B_n},\mathbf{P}_{B_n}\): complex steerable decomposition of label and prediction image at level \(n\); \(I(\mathbf{Y}_{B_n}, \mathbf{Y}_{B_n})\): mutual information between \(\mathbf{Y}_{B_n}\) and \(\mathbf{P}_{B_n}\)}
    \label{fig:schematic}
\end{figure*}

\vspace{-7pt}
\subsection{Loss Functions for Semantic Segmentation}
While the architectural advancements in segmentation models have been significant, the performance of these networks is highly influenced by the design of the loss functions. Pixel-wise cross entropy loss (CE) minimizes the log likelihood of the prediction error but is significantly prone to class imbalance. To address this issue, class balanced cross entropy (BCE), proposed by \cite{long2015fully}, employs higher weights for classes with fewer pixels. Focal loss assigns higher weights to misclassified pixels with high probabilities \cite{ross2017focal}. Region-based losses, including Dice loss \cite{milletari2016v}, Tversky loss \cite{salehi2017tversky}, and Jaccard loss \cite{rahman2016optimizing}, inherently handle class imbalance but fail to address instance imbalance within the same class, such as thin boundaries and small objects.

To tackle instance imbalance, weighted loss functions have been proposed. In the original U-Net paper \cite{ronneberger2015u}, the weighted cross entropy (WCE) was introduced, employing a distance-based weight map to emphasize thin boundaries between objects. However, WCE assigns weights only to boundary pixels, neglecting object pixels. The adaptive boundary weighted (ABW) loss \cite{liu2022boundary} extends this approach by applying distance-based weights to both boundary and object pixels, while the Skea-Topo loss further improved the weight map based on boundary and object skeletons \cite{liu2024enhancing}. Despite their contributions, weighted losses suffer from two major limitations: (1) the weight maps are precomputed and fixed, failing to adapt to errors during training, and (2) they often generate thicker boundaries, which preserve topology but compromise metrics like Dice score and mIoU, as observed in our qualitative results.

Several methods address instance imbalance dynamically during training, but at the cost of computational efficiency. Topology-based approaches, such as persistent homology methods \cite{stucki2023topologically,oner2023persistent}, describe image topologies and identify critical pixels but are computationally expensive, with cubic complexity to image size. The clDice loss \cite{shit2021cldice} employs a soft skeletonization algorithm to detect topological errors, primarily focusing on thin-boundary objects like retinal blood vessels. Similarly, Boundary Loss \cite{kervadec2019boundary} and Hausdorff Distance Loss \cite{karimi2019reducing} refine boundaries but incur significant computational overhead. Region Mutual Information (RMI) loss \cite{zhao2019region} captures pixel interdependencies over regions, but struggles with scalability for large-scale regional analysis due to the high computation cost. These losses either prioritize small regions at the expense of global accuracy or require extensive computational resources, necessitating more efficient and balanced approaches.

\vspace{-7pt}
\subsection{Wavelet-Based Loss Functions}

Wavelet-based metrics were first introduced as Complex Wavelet Structural Similarity (CW-SSIM) \cite{sampat2009complex}, known for their robustness to small translations and rotations. In the deep learning era, wavelet-based methods have been employed in loss functions, leveraging their ability to analyze multiscale and multiresolution features. These methods have shown promise in tasks like sketch-to-image translation \cite{kim2023whfl}, image super-resolution \cite{korkmaz2024training}, image dehazing \cite{yang2020net}, and material analysis \cite{prantl2022wavelet}. However, to the best of our knowledge, wavelet-based loss functions have yet to be explored in semantic segmentation.

Existing wavelet-based loss functions typically rely on $L_1$ \cite{zhu2021wavelet,korkmaz2024training,prantl2022wavelet} or $L_2$ distances \cite{kim2023whfl}, or structural similarity (SSIM) \cite{yang2020net} in the decomposed domain. While effective, these methods are less suited for handling high-dimensional data with complex directional features, as in wavelet transforms, and may be vulnerable to noise. The proposed CWMI loss leverages mutual information between wavelet-based subband images, effectively capturing multiscale dependencies. As demonstrated in later ablation tests, CWMI outperforms traditional metrics like $L_1$, $L_2$, and SSIM, offering superior segmentation performance and robustness.

\vspace{-7pt}
\section{Methods}

To compute the proposed Complex Wavelet Mutual Information (CWMI) loss, the prediction and ground truth label matrices are first decomposed into subbands using the complex steerable pyramid. Mutual information (MI) is then calculated for each subband. The CWMI loss is defined as the sum of the MI values across all subbands, as illustrated in Figure \ref{fig:schematic}.

\vspace{-7pt}
\subsection{Complex Wavelet Decomposition}
\paragraph{Wavelet Transform}
Wavelet transforms are widely used for multiscale analysis, enabling decomposition of an image into frequency subbands while preserving spatial information. Unlike traditional Fourier transforms, which analyze global frequency components, wavelet transforms provide a localized frequency representation, making them well-suited for tasks involving spatially-varying structures such as semantic segmentation. For an image \(I(x, y)\), the wavelet transform is defined as:
\begin{equation}
W_{\psi}(s, t) = \int \int I(x, y) \psi_{s, t}(x, y) \,dx\,dy
\end{equation}
where \( \psi_{s, t}(x, y) \) is a scaled and translated version of the mother wavelet function \( \psi(x, y) \), with \( s \) controlling the scale and \( t \) the translation.

For discrete signals, the Discrete Wavelet Transform (DWT) decomposes an image into progressively lower resolution subbands using filter banks. However, traditional wavelet decompositions suffer from limited orientation selectivity, capturing only fixed horizontal, vertical, and diagonal directions.

\begin{figure}[!h]
    \centering
    \includegraphics[width=0.9\linewidth]{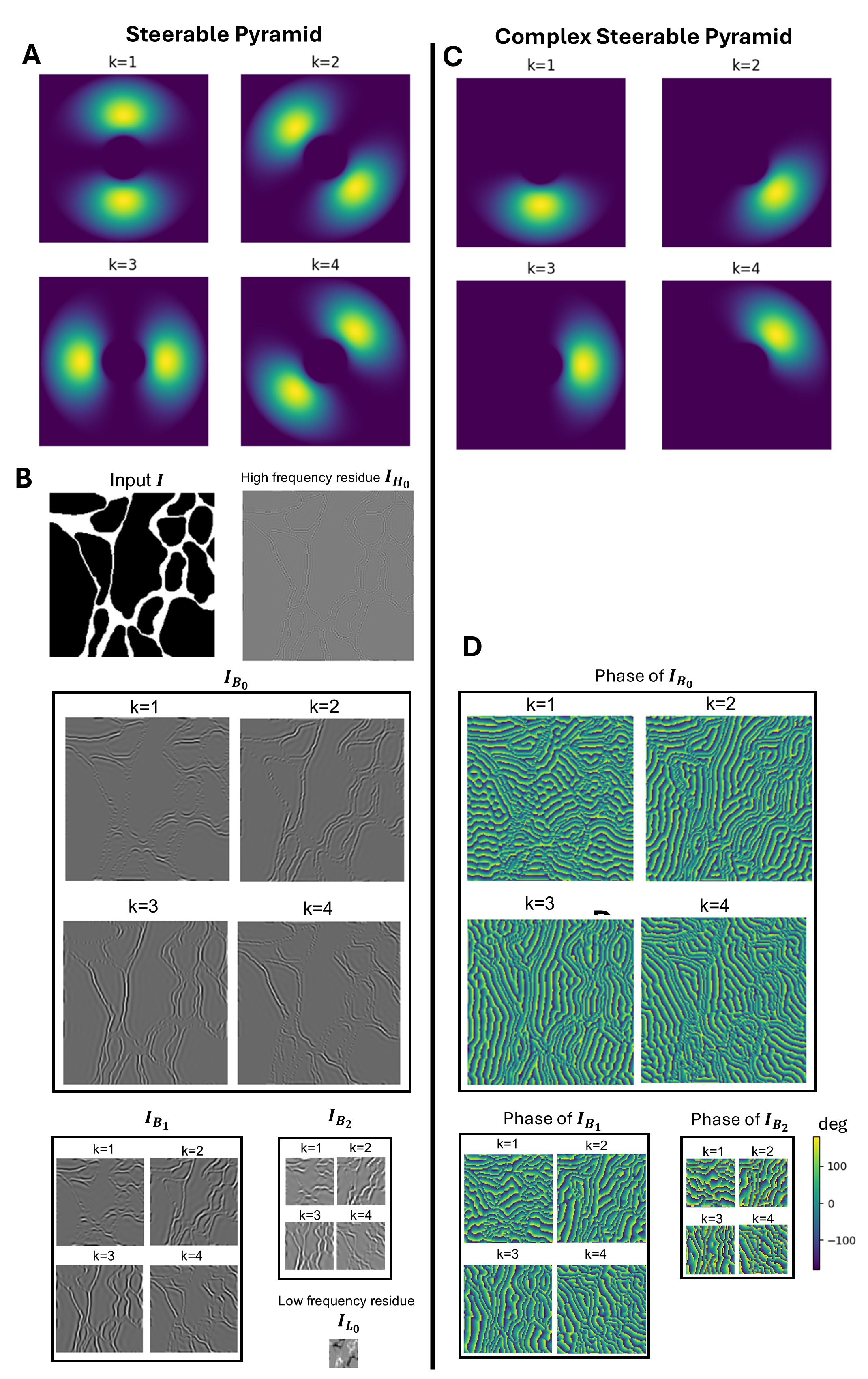}
    \vspace{-10pt}
    \caption{Steerable pyramid and complex steerable pyramid. (A) Orientation-selective band-pass filters of the steerable pyramid. (B) Example decomposition using a steerable pyramid with N=3, K=4. (C) Band-pass filters of the complex steerable pyramid, where negative frequency components are discarded. (D) Phase representation of the complex steerable pyramid output, with the real part identical to that of the steerable pyramid.}
    \label{fig:2}
\end{figure}

\vspace{-10pt}
\paragraph{Steerable Pyramid}
To overcome these limitations, steerable pyramid extends the wavelet framework by introducing orientation-sensitive band-pass filters, significantly enhancing orientation selectivity. Unlike DWT, which provides a non-redundant representation, the steerable pyramid offers a flexible, redundant image representation, facilitating improved multiscale analysis. This decomposition is achieved through the iterative application of steerable band-pass filters followed by downsampling. In the frequency domain, the band-pass filter for the \(k_{th}\) orientation is expressed in polar coordinates \((r, \theta)\) as:
\begin{equation}
B_k(r, \theta) = H(r) G_k(\theta), \quad k \in [1, K],
\end{equation}
where \(H(r)\) and \(G_k(\theta)\) represent the radial and angular components, respectively:
\begin{equation}
H(r) =
\begin{cases}
    \cos\left(\frac{\pi}{2} \log_2 \left(\frac{2r}{\pi}\right)\right), & \frac{\pi}{4} < r < \frac{\pi}{2}, \\
    1, & r \leq \frac{\pi}{2}, \\
    0, & r \geq \frac{\pi}{4}.
\end{cases}
\end{equation}
\begin{equation}\label{G real}
G_k(\theta) =
    \alpha_k \left|\cos\left(\theta - \frac{\pi k}{K}\right)\right|^{K-1},
\end{equation}
where \(K\) is the number of orientations and \(\alpha_k = 2^{k-1} \frac{(K-1)!}{\sqrt{K[2(K-1)]!}}\). Figure \ref{fig:2}A provides an illustration of the band filter with \(K=4\).

With total recursive levels \(N\), an image \(I\) can be decomposed as:
\begin{equation}\label{eq:7}
\mathbf{I} \rightarrow
\left|
\begin{array}{ll}
    \mathbf{I}_{H_0} \in \mathbf{R}^{H_0 \times W_0}, \\
    \mathbf{I}_{B_1} \in \mathbf{R}^{K \times H_0 \times W_0}, \\
    \mathbf{I}_{B_2} \in \mathbf{R}^{K \times H_1 \times W_1}, \\
    \dots \\
    \mathbf{I}_{B_N} \in \mathbf{R}^{K \times H_{N-1} \times W_{N-1}}, \\
    \mathbf{I}_{L_0} \in \mathbf{R}^{H_{N-1} \times W_{N-1}},
\end{array}
\right.
\end{equation}
where \(\mathbf{I}_{H_0}\) and \(\mathbf{I}_{L_0}\) are high-frequency and low-frequency residues, and \(\mathbf{I}_{B_n}\) represents subband images at level \(n\) with \(k\)-th direction concatenated in the first dimension. Figure \ref{fig:2}B shows an example of the decomposed output of an input image \(\mathbf{I}\).

\vspace{-10pt}
\paragraph{Complex Steerable Pyramid}
Although the steerable pyramid effectively captures amplitude information across multiple orientations, it lacks the ability to extract local phase information, which is crucial for encoding structural features such as edges and corners \cite{canny1986computational}. To address this limitation, \cite{portilla2000parametric} introduced the complex steerable pyramid, which extends the conventional steerable pyramid by converting its decomposed images into their analytic signal representation. In this formulation, the real part remains unchanged, while the imaginary part is obtained via the Hilbert transform of the real component. In the Fourier domain, this transformation is equivalent to discarding negative frequency components, as illustrated in Figure \ref{fig:2}C.

For the complex steerable pyramid, the angular component \(G_k(\theta)\) is modified as:
\begin{equation}\label{G complex}
\tilde{G_k}(\theta) =
\begin{cases}
    2\alpha_k \left[\cos\left(\theta - \frac{\pi k}{K}\right)\right]^{K-1}, & \left| \theta - \frac{\pi k}{K} \right| < \frac{\pi}{2}, \\
    0, & \text{otherwise}.
\end{cases}
\end{equation}
This modification enables image decomposition into complex subbands, where phase information encodes critical structural features such as edges and corners (Figure \ref{fig:2}D), while amplitude represents feature strength.

\vspace{-5pt}
\subsection{Mutual Information in Complex Wavelet Domain}

According to Equation \ref{eq:7}, the ground truth \(\mathbf{Y}\) and the prediction \(\mathbf{P}\) are decomposed into subbands \(\mathbf{Y}_{B_n}\) and \(\mathbf{P}_{B_n} \in \mathbf{R}^{K \times H_{n-1} \times W_{n-1}}\) for each level \(n \in [1, N]\). For each pixel \((x, y)\) at level \(n\), the \(K\)-directional features are treated as \(K\)-dimensional random variables. 

For mutual information approximation, several studies, such as MINE \cite{belghazi2018mutual} and Deep InfoMax \cite{hjelm2018learning}, employ neural network-based estimators to produce tight lower bounds. While these methods are theoretically well-founded, they introduce additional training overhead, which increases algorithmic complexity and may hinder the efficiency of the loss function. Therefore, in this work, we adopt the mutual information estimation approach proposed by \cite{zhao2019region}:
\begin{equation}
I_l(\mathbf{Y}_{B_n}; \mathbf{P}_{B_n}) \approx -\frac{1}{2}\log\det(\mathbf{M}_n),
\end{equation}
\begin{equation}\label{eq:9}
\mathbf{M}_n = \mathbf{\Sigma}_{\mathbf{Y}_{B_n}} - \text{Cov}(\mathbf{Y}_{B_n}, \mathbf{P}_{B_n}) (\mathbf{\Sigma}_{\mathbf{P}_{B_n}}^{-1})^T \text{Cov}(\mathbf{Y}_{B_n}, \mathbf{P}_{B_n})^T
\end{equation}
where \(\mathbf{\Sigma}_{\mathbf{Y}_{B_n}}\) and \(\mathbf{\Sigma}_{\mathbf{P}_{B_n}}\) are covariance matrices, and \(\text{Cov}(\mathbf{Y}_{B_n}, \mathbf{P}_{B_n})\) is the cross-covariance. 

When the ground truth and predictions are decomposed into complex subbands, this equation is extended with Hermitian transpose \(H\):
\begin{equation}\label{eq:10}
\mathbf{\tilde{M}}_n = \mathbf{\Sigma}_{\mathbf{Y}_{B_n}} - \text{Cov}(\mathbf{Y}_{B_n}, \mathbf{P}_{B_n}) (\mathbf{\Sigma}_{\mathbf{P}_{B_n}}^{-1})^H \text{Cov}(\mathbf{Y}_{B_n}, \mathbf{P}_{B_n})^H
\end{equation}
where the covariance and cross-covariance are calculated as:
\begin{equation}\label{eq:11}
    \mathbf{\Sigma}_{\mathbf{Y}_{B_n}}=E[(\mathbf{Y}_{B_n} - E[\mathbf{Y}_{B_n}])(\mathbf{Y}_{B_n} - E[\mathbf{Y}_{B_n}])^H]
\end{equation}
\begin{equation}\label{eq:12}
    \mathbf{\Sigma}_{\mathbf{P}_{B_n}}=E[(\mathbf{P}_{B_n} - E[\mathbf{P}_{B_n}])(\mathbf{P}_{B_n} - E[\mathbf{P}_{B_n}])^H]
\end{equation}
\begin{equation}
    \text{Cov}(\mathbf{Y}_{B_n}, \mathbf{P}_{B_n})=E[(\mathbf{Y}_{B_n} - E[\mathbf{Y}_{B_n}])(\mathbf{P}_{B_n} - E[\mathbf{P}_{B_n}])^H]
\end{equation}

Finally, the CWMI loss is computed as the sum of MI across all levels, which combines with cross entropy loss to integrate pixel-wise loss:
\begin{equation}\label{eq:13}
CWMI(\mathbf{Y}, \mathbf{P}) = (1-\lambda)\sum_{n=1}^N -I_l(\mathbf{Y}_{B_n}; \mathbf{P}_{B_n})+\lambda L_{ce}(\mathbf{Y}, \mathbf{P}).
\end{equation}

\begin{table*}\label{Table 1}
\scriptsize
    \centering
    \setlength{\tabcolsep}{1pt}
    \begin{tabular}{lccccc |ccccc}
        \Xhline{1pt}
        \multicolumn{11}{c}{\textbf{SENMI3D}}\\
        \hline
        ~&~&~&UNet&~&~&
        ~&~&AttenUNet&~ &~\\
Methods  & 
mIoU$\uparrow$ & mDice$\uparrow$ & VI$\downarrow$ & ARI$\uparrow$ & HD$\downarrow$ & 
mIoU$\uparrow$ & mDice$\uparrow$ & VI$\downarrow$ & ARI$\uparrow$ & HD$\downarrow$\\

\hline

CE	&\(.753_{\pm.005}\)	&\(.851_{\pm.004}\)	&\(2.082_{\pm.270}\)	&\(.500_{\pm.039}\)	&\(1.143_{\pm.176}\)	&\
\(.751_{\pm.009}\)	&\(.850_{\pm.007}\)	&\(1.914_{\pm.361}\)	&\(.525_{\pm.056}\)	&\(1.178_{\pm.261}\)	\\
BCE	&\(.736_{\pm.006}\)	&\(.841_{\pm.004}\)	&\(1.756_{\pm.156}\)	&\(.561_{\pm.020}\)	&\(1.413_{\pm.088}\)	&\
\(.735_{\pm.010}\)	&\(.840_{\pm.007}\)	&\(1.835_{\pm.150}\)	&\(.553_{\pm.023}\)	&\(1.417_{\pm.224}\)	\\
Dice	&\(.767_{\pm.004}\)	&\(.862_{\pm.003}\)	&\(1.406_{\pm.085}\)	&\(.604_{\pm.017}\)	&\(.919_{\pm.154}\)	&\
\(.768_{\pm.005}\)	&\(.862_{\pm.004}\)	&\(1.507_{\pm.086}\)	&\(.583_{\pm.016}\)	&\(\mathbf{.804_{\pm.072}}\)	\\
Focal	&\(.728_{\pm.007}\)	&\(.835_{\pm.005}\)	&\(1.751_{\pm.078}\)	&\(.556_{\pm.016}\)	&\(1.725_{\pm.225}\)	&\
\(.725_{\pm.005}\)	&\(.833_{\pm.003}\)	&\(1.911_{\pm.189}\)	&\(.540_{\pm.019}\)	&\(1.803_{\pm.107}\)	\\
Jaccard	&\(.766_{\pm.004}\)	&\(.861_{\pm.003}\)	&\(1.357_{\pm.114}\)	&\(.614_{\pm.019}\)	&\(.904_{\pm.087}\)	&\
\(.762_{\pm.003}\)	&\(.858_{\pm.002}\)	&\(1.360_{\pm.079}\)	&\(.609_{\pm.007}\)	&\(1.000_{\pm.125}\)	\\
Tversky	&\(.765_{\pm.004}\)	&\(.861_{\pm.003}\)	&\(1.328_{\pm.077}\)	&\(.617_{\pm.017}\)	&\(.990_{\pm.183}\)	&\
\(.765_{\pm.003}\)	&\(.860_{\pm.002}\)	&\(1.303_{\pm.057}\)	&\(.621_{\pm.009}\)	&\(.969_{\pm.101}\)	\\
WCE	&\(.714_{\pm.005}\)	&\(.825_{\pm.003}\)	&\(1.987_{\pm.398}\)	&\(.525_{\pm.042}\)	&\(1.575_{\pm.250}\)	&\
\(.713_{\pm.005}\)	&\(.825_{\pm.004}\)	&\(1.818_{\pm.034}\)	&\(.543_{\pm.009}\)	&\(1.613_{\pm.192}\)	\\
ABW	&\(.605_{\pm.066}\)	&\(.739_{\pm.057}\)	&\(3.637_{\pm1.703}\)	&\(.299_{\pm.187}\)	&\(4.605_{\pm2.926}\)	&\
\(.616_{\pm.072}\)	&\(.749_{\pm.063}\)	&\(3.077_{\pm1.303}\)	&\(.366_{\pm.142}\)	&\(4.045_{\pm3.220}\)	\\
Skea-topo	&\(.572_{\pm.149}\)	&\(.671_{\pm.197}\)	&\(3.822_{\pm2.560}\)	&\(.300_{\pm.259}\)	&\(1.618_{\pm.207}\)	&\
\(.602_{\pm.095}\)	&\(.729_{\pm.096}\)	&\(3.759_{\pm2.324}\)	&\(.306_{\pm.243}\)	&\(4.889_{\pm5.928}\)	\\
RMI	&\(.764_{\pm.006}\)	&\(.859_{\pm.005}\)	&\(1.443_{\pm.207}\)	&\(.587_{\pm.037}\)	&\(.943_{\pm.130}\)	&\
\(.764_{\pm.008}\)	&\(.859_{\pm.006}\)	&\(1.404_{\pm.124}\)	&\(.585_{\pm.028}\)	&\(.958_{\pm.170}\)	\\
clDice	&\(.706_{\pm.014}\)	&\(.819_{\pm.011}\)	&\(2.049_{\pm.351}\)	&\(.502_{\pm.052}\)	&\(1.489_{\pm.218}\)	&\
\(.701_{\pm.022}\)	&\(.815_{\pm.016}\)	&\(1.887_{\pm.161}\)	&\(.518_{\pm.034}\)	&\(1.720_{\pm.244}\)	\\
Sensitive	&\(.763_{\pm.006}\)	&\(.858_{\pm.005}\)	&\(1.543_{\pm.069}\)	&\(.576_{\pm.023}\)	&\(.948_{\pm.143}\)	&\
\(.757_{\pm.006}\)	&\(.854_{\pm.004}\)	&\(1.822_{\pm.313}\)	&\(.527_{\pm.049}\)	&\(1.016_{\pm.139}\)	\\

\hline

CWMI-Real	&\(\underline{.776_{\pm.004}}^{**}\)	&\(\underline{.867_{\pm.003}}^{**}\)	&\(\underline{1.205_{\pm.059}}^{**}\)	&\(\underline{.634_{\pm.013}}^{*}\)	&\(\underline{.807_{\pm.117}}^{*}\)	&\
\(\underline{.775_{\pm.004}}^{**}\)	&\(\underline{.867_{\pm.003}}^{**}\)	&\(\underline{1.193_{\pm.070}}^{**}\)	&\(\underline{.634_{\pm.014}}^{*}\)	&\(.824_{\pm.103}\)	\\
CWMI	&\(\mathbf{.778_{\pm.004}}^{***}\)	&\(\mathbf{.869_{\pm.003}}^{***}\)	&\(\mathbf{1.162_{\pm.068}}^{***}\)	&\(\mathbf{.638_{\pm.015}}^{*}\)	&\(\mathbf{.739_{\pm.095}}^{**}\)	&\
\(\mathbf{.777_{\pm.006}}^{**}\)	&\(\mathbf{.868_{\pm.005}}^{**}\)	&\(\mathbf{1.162_{\pm.085}}^{**}\)	&\(\mathbf{.639_{\pm.019}}^{*}\)	&\(\underline{.807_{\pm.069}}\)	\\

        \Xhline{1pt}
        \multicolumn{11}{c}{\textbf{GlaS}}\\
        \hline
        ~&~&~&UNet&~&~&
        ~&~&AttenUNet&~ &~\\
Methods  & 
mIoU$\uparrow$ & mDice$\uparrow$ & VI$\downarrow$ & ARI$\uparrow$ & HD$\downarrow$ & 
mIoU$\uparrow$ & mDice$\uparrow$ & VI$\downarrow$ & ARI$\uparrow$ & HD$\downarrow$\\
        \hline
CE	&\(.637_{\pm.204}\)	&\(.724_{\pm.192}\)	&\(.933_{\pm.043}\)	&\(.400_{\pm.333}\)	&\(6.633_{\pm4.654}\)	&\
\(.640_{\pm.209}\)	&\(.725_{\pm.195}\)	&\(.906_{\pm.107}\)	&\(.406_{\pm.327}\)	&\(7.914_{\pm6.637}\)	\\
BCE	&\(.614_{\pm.226}\)	&\(.702_{\pm.213}\)	&\(1.032_{\pm.152}\)	&\(.355_{\pm.355}\)	&\(4.165_{\pm1.929}\)	&\
\(.619_{\pm.222}\)	&\(.708_{\pm.208}\)	&\(.986_{\pm.139}\)	&\(.365_{\pm.358}\)	&\(4.354_{\pm1.869}\)	\\
Dice	&\(.632_{\pm.201}\)	&\(.719_{\pm.189}\)	&\(.887_{\pm.080}\)	&\(.396_{\pm.322}\)	&\(7.004_{\pm4.670}\)	&\
\(.636_{\pm.197}\)	&\(.723_{\pm.187}\)	&\(.954_{\pm.056}\)	&\(.396_{\pm.322}\)	&\(7.335_{\pm4.843}\)	\\
Focal	&\(.581_{\pm.253}\)	&\(.672_{\pm.241}\)	&\(1.093_{\pm.103}\)	&\(.330_{\pm.356}\)	&\(4.447_{\pm2.023}\)	&\
\(.581_{\pm.252}\)	&\(.672_{\pm.240}\)	&\(1.043_{\pm.109}\)	&\(.336_{\pm.369}\)	&\(5.040_{\pm2.802}\)	\\
Jaccard	&\(.637_{\pm.205}\)	&\(.723_{\pm.193}\)	&\(.947_{\pm.076}\)	&\(.397_{\pm.335}\)	&\(6.979_{\pm4.511}\)	&\
\(.638_{\pm.202}\)	&\(.725_{\pm.190}\)	&\(.888_{\pm.076}\)	&\(.406_{\pm.331}\)	&\(6.758_{\pm4.431}\)	\\
Tversky	&\(.634_{\pm.205}\)	&\(.722_{\pm.193}\)	&\(.953_{\pm.031}\)	&\(.394_{\pm.332}\)	&\(6.705_{\pm4.398}\)	&\
\(.627_{\pm.195}\)	&\(.717_{\pm.186}\)	&\(.924_{\pm.102}\)	&\(.384_{\pm.314}\)	&\(7.521_{\pm4.493}\)	\\
WCE	&\(.829_{\pm.010}\)	&\(.902_{\pm.007}\)	&\(.832_{\pm.075}\)	&\(.721_{\pm.008}\)	&\(\underline{2.264_{\pm.210}}\)	&\
\(.823_{\pm.016}\)	&\(.898_{\pm.011}\)	&\(.873_{\pm.097}\)	&\(.705_{\pm.023}\)	&\(\underline{2.590_{\pm.602}}\)	\\
ABW	&\(.760_{\pm.012}\)	&\(.857_{\pm.008}\)	&\(1.468_{\pm.079}\)	&\(.619_{\pm.027}\)	&\(3.527_{\pm.508}\)	&\
\(.762_{\pm.019}\)	&\(.858_{\pm.013}\)	&\(1.472_{\pm.103}\)	&\(.614_{\pm.037}\)	&\(3.623_{\pm.887}\)	\\
Skea-topo	&\(.788_{\pm.008}\)	&\(.876_{\pm.005}\)	&\(1.259_{\pm.042}\)	&\(.658_{\pm.002}\)	&\(2.985_{\pm.765}\)	&\
\(.784_{\pm.016}\)	&\(.873_{\pm.010}\)	&\(1.299_{\pm.086}\)	&\(.645_{\pm.022}\)	&\(3.363_{\pm.609}\)	\\
RMI	&\(\underline{.839_{\pm.011}}\)	&\(\underline{.907_{\pm.007}}\)	&\(.820_{\pm.082}\)	&\(.726_{\pm.024}\)	&\(2.644_{\pm.372}\)	&\
\(.835_{\pm.015}\)	&\(.905_{\pm.010}\)	&\(.822_{\pm.083}\)	&\(.720_{\pm.039}\)	&\(2.701_{\pm.366}\)	\\
clDice	&\(.816_{\pm.017}\)	&\(.894_{\pm.012}\)	&\(.920_{\pm.107}\)	&\(.698_{\pm.026}\)	&\(2.809_{\pm.476}\)	&\
\(.800_{\pm.013}\)	&\(.884_{\pm.009}\)	&\(1.004_{\pm.113}\)	&\(.684_{\pm.026}\)	&\(3.146_{\pm.286}\)	\\
Sensitive	&\(.821_{\pm.014}\)	&\(.897_{\pm.009}\)	&\(.941_{\pm.101}\)	&\(.694_{\pm.033}\)	&\(2.916_{\pm.394}\)	&\
\(.824_{\pm.018}\)	&\(.899_{\pm.011}\)	&\(.880_{\pm.108}\)	&\(.702_{\pm.041}\)	&\(2.757_{\pm.398}\)	\\

\hline

CWMI-Real	&\(.838_{\pm.014}\)	&\(.907_{\pm.010}\)	&\(\underline{.798_{\pm.068}}\)	&\(\underline{.727_{\pm.027}}\)	&\(2.813_{\pm.653}\)	&\
\(\underline{.842_{\pm.022}}\)	&\(\underline{.909_{\pm.015}}\)	&\(\underline{.788_{\pm.063}}\)	&\(\underline{.724_{\pm.038}}\)	&\(2.758_{\pm.764}\)	\\
CWMI	&\(\mathbf{.843_{\pm.016}}\)	&\(\mathbf{.910_{\pm.011}}\)	&\(\mathbf{.761_{\pm.081}}^{*}\)	&\(\mathbf{.735_{\pm.026}}\)	&\(\mathbf{2.569_{\pm.466}}\)	&\
\(\mathbf{.844_{\pm.007}}\)	&\(\mathbf{.911_{\pm.005}}\)	&\(\mathbf{.755_{\pm.106}}\)	&\(\mathbf{.737_{\pm.015}}\)	&\(\mathbf{2.569_{\pm.547}}\)	\\

        \Xhline{1pt}
        \multicolumn{11}{c}{\textbf{DRIVE}}\\
        \hline
        ~&~&~&UNet&~&~&
        ~&~&AttenUNet&~ &~\\
Methods  & 
mIoU$\uparrow$ & mDice$\uparrow$ & VI$\downarrow$ & ARI$\uparrow$ & HD$\downarrow$ & 
mIoU$\uparrow$ & mDice$\uparrow$ & VI$\downarrow$ & ARI$\uparrow$ & HD$\downarrow$\\
        \hline
CE	&\(.770_{\pm.014}\)	&\(.856_{\pm.011}\)	&\(1.379_{\pm.136}\)	&\(.406_{\pm.059}\)	&\(2.344_{\pm.398}\)	&\
\(.757_{\pm.017}\)	&\(.845_{\pm.014}\)	&\(1.436_{\pm.126}\)	&\(.372_{\pm.059}\)	&\(2.932_{\pm.875}\)	\\
BCE	&\(.742_{\pm.004}\)	&\(.835_{\pm.003}\)	&\(1.501_{\pm.175}\)	&\(.399_{\pm.088}\)	&\(2.355_{\pm.434}\)	&\
\(.746_{\pm.012}\)	&\(.838_{\pm.010}\)	&\(1.443_{\pm.106}\)	&\(.427_{\pm.033}\)	&\(2.169_{\pm.264}\)	\\
Dice	&\(.779_{\pm.018}\)	&\(.863_{\pm.014}\)	&\(1.293_{\pm.116}\)	&\(.471_{\pm.045}\)	&\(1.696_{\pm.275}\)	&\
\(.776_{\pm.015}\)	&\(.861_{\pm.012}\)	&\(1.335_{\pm.136}\)	&\(.445_{\pm.079}\)	&\(2.146_{\pm.832}\)	\\
Focal	&\(.737_{\pm.008}\)	&\(.832_{\pm.006}\)	&\(1.404_{\pm.168}\)	&\(.471_{\pm.113}\)	&\(2.147_{\pm.331}\)	&\
\(.745_{\pm.011}\)	&\(.837_{\pm.009}\)	&\(1.410_{\pm.139}\)	&\(.450_{\pm.075}\)	&\(2.141_{\pm.400}\)	\\
Jaccard	&\(.761_{\pm.013}\)	&\(.850_{\pm.011}\)	&\(1.289_{\pm.181}\)	&\(.502_{\pm.088}\)	&\(1.988_{\pm.641}\)	&\
\(.766_{\pm.020}\)	&\(.854_{\pm.015}\)	&\(1.271_{\pm.145}\)	&\(.521_{\pm.051}\)	&\(1.623_{\pm.313}\)	\\
Tversky	&\(.752_{\pm.020}\)	&\(.843_{\pm.016}\)	&\(1.309_{\pm.086}\)	&\(.517_{\pm.052}\)	&\(1.638_{\pm.369}\)	&\
\(.768_{\pm.014}\)	&\(.856_{\pm.011}\)	&\(1.265_{\pm.129}\)	&\(.521_{\pm.048}\)	&\(1.702_{\pm.298}\)	\\
WCE	&\(.741_{\pm.013}\)	&\(.835_{\pm.011}\)	&\(1.417_{\pm.069}\)	&\(.462_{\pm.019}\)	&\(2.100_{\pm.498}\)	&\
\(.733_{\pm.009}\)	&\(.828_{\pm.007}\)	&\(1.441_{\pm.031}\)	&\(.444_{\pm.039}\)	&\(2.326_{\pm.848}\)	\\
ABW	&\(-\)	&\(-\)	&\(-\)	&\(-\)	&\(-\)	&\
\(-\)	&\(-\)	&\(-\)	&\(-\)	&\(-\)	\\
Skea-topo	&\(-\)	&\(-\)	&\(-\)	&\(-\)	&\(-\)	&\
\(.599_{\pm.121}\)	&\(.674_{\pm.169}\)	&\(1.602_{\pm.113}\)	&\(.265_{\pm.250}\)	&\(5.791_{\pm4.492}\)	\\
RMI	&\(.787_{\pm.012}\)	&\(.869_{\pm.010}\)	&\(1.282_{\pm.127}\)	&\(.449_{\pm.064}\)	&\(1.797_{\pm.489}\)	&\
\(.785_{\pm.009}\)	&\(.867_{\pm.007}\)	&\(1.279_{\pm.096}\)	&\(.448_{\pm.047}\)	&\(1.884_{\pm.340}\)	\\
clDice	&\(.482_{\pm.094}\)	&\(.598_{\pm.085}\)	&\(2.010_{\pm.119}\)	&\(.207_{\pm.129}\)	&\(9.813_{\pm8.153}\)	&\
\(.509_{\pm.063}\)	&\(.617_{\pm.066}\)	&\(1.912_{\pm.157}\)	&\(.238_{\pm.161}\)	&\(15.704_{\pm11.080}\)	\\
Sensitive	&\(.763_{\pm.008}\)	&\(.850_{\pm.007}\)	&\(1.440_{\pm.060}\)	&\(.376_{\pm.039}\)	&\(2.824_{\pm.562}\)	&\
\(.757_{\pm.018}\)	&\(.846_{\pm.015}\)	&\(1.444_{\pm.040}\)	&\(.380_{\pm.028}\)	&\(3.134_{\pm.782}\)	\\

\hline

CWMI-Real	&\(\underline{.788_{\pm.016}}\)	&\(\underline{.870_{\pm.012}}\)	&\(\underline{1.104_{\pm.175}}^{*}\)	&\(\underline{.582_{\pm.075}}\)	&\(\underline{1.309_{\pm.299}}^{*}\)	&\
\(\underline{.787_{\pm.026}}\)	&\(\underline{.870_{\pm.019}}\)	&\(\underline{1.072_{\pm.113}}^{**}\)	&\(\underline{.594_{\pm.057}}^{*}\)	&\(\underline{1.398_{\pm.468}}\)	\\
CWMI	&\(\mathbf{.798_{\pm.012}}^{*}\)	&\(\mathbf{.878_{\pm.009}}^{*}\)	&\(\mathbf{1.032_{\pm.176}}^{*}\)	&\(\mathbf{.613_{\pm.082}}^{*}\)	&\(\mathbf{1.079_{\pm.382}}^{*}\)	&\
\(\mathbf{.795_{\pm.015}}^{*}\)	&\(\mathbf{.875_{\pm.012}}^{*}\)	&\(\mathbf{1.007_{\pm.162}}^{**}\)	&\(\mathbf{.622_{\pm.062}}^{**}\)	&\(\mathbf{1.219_{\pm.410}}^{*}\)	\\

\Xhline{1pt}
\multicolumn{11}{c}{\textbf{MASS ROAD}}\\
        \hline
        ~&~&~&UNet&~&~&
        ~&~&AttenUNet&~ &~\\
Methods  & 
mIoU$\uparrow$ & mDice$\uparrow$ & VI$\downarrow$ & ARI$\uparrow$ & HD$\downarrow$ & 
mIoU$\uparrow$ & mDice$\uparrow$ & VI$\downarrow$ & ARI$\uparrow$ & HD$\downarrow$\\
        \hline
CE	&\(.733_{\pm.015}\)	&\(.826_{\pm.013}\)	&\(3.473_{\pm.608}\)	&\(.194_{\pm.083}\)	&\(1.927_{\pm3.301}\)	&\
\(.640_{\pm.209}\)	&\(.725_{\pm.195}\)	&\(.906_{\pm.107}\)	&\(.406_{\pm.327}\)	&\(7.914_{\pm6.637}\)	\\
BCE	&\(.689_{\pm.011}\)	&\(.794_{\pm.009}\)	&\(1.708_{\pm.265}\)	&\(.560_{\pm.059}\)	&\(12.792_{\pm3.684}\)	&\
\(.619_{\pm.222}\)	&\(.708_{\pm.208}\)	&\(.986_{\pm.139}\)	&\(.365_{\pm.358}\)	&\(4.354_{\pm1.869}\)	\\
Dice	&\(.761_{\pm.011}\)	&\(.849_{\pm.009}\)	&\(1.601_{\pm.138}\)	&\(.558_{\pm.042}\)	&\(11.189_{\pm4.532}\)	&\
\(.636_{\pm.197}\)	&\(.723_{\pm.187}\)	&\(.954_{\pm.056}\)	&\(.396_{\pm.322}\)	&\(7.335_{\pm4.843}\)	\\
Focal	&\(.671_{\pm.015}\)	&\(.780_{\pm.013}\)	&\(1.717_{\pm.150}\)	&\(.556_{\pm.042}\)	&\(11.576_{\pm1.950}\)	&\
\(.581_{\pm.252}\)	&\(.672_{\pm.240}\)	&\(1.043_{\pm.109}\)	&\(.336_{\pm.369}\)	&\(5.040_{\pm2.802}\)	\\
Jaccard	&\(.759_{\pm.008}\)	&\(.848_{\pm.006}\)	&\(1.332_{\pm.041}\)	&\(.630_{\pm.008}\)	&\(11.274_{\pm2.805}\)	&\
\(.638_{\pm.202}\)	&\(.725_{\pm.190}\)	&\(.888_{\pm.076}\)	&\(.406_{\pm.331}\)	&\(6.758_{\pm4.431}\)	\\
Tversky	&\(.755_{\pm.009}\)	&\(.845_{\pm.007}\)	&\(1.379_{\pm.196}\)	&\(.618_{\pm.061}\)	&\(11.501_{\pm4.470}\)	&\
\(.627_{\pm.195}\)	&\(.717_{\pm.186}\)	&\(.924_{\pm.102}\)	&\(.384_{\pm.314}\)	&\(7.521_{\pm4.493}\)	\\
WCE	&\(.650_{\pm.016}\)	&\(.762_{\pm.014}\)	&\(1.726_{\pm.103}\)	&\(.556_{\pm.004}\)	&\(1.845_{\pm.470}\)	&\
\(.823_{\pm.016}\)	&\(.898_{\pm.011}\)	&\(.873_{\pm.097}\)	&\(.705_{\pm.023}\)	&\(2.590_{\pm.602}\)	\\
ABW	&\(.685_{\pm.014}\)	&\(.791_{\pm.012}\)	&\(1.678_{\pm.124}\)	&\(.566_{\pm.030}\)	&\(11.654_{\pm4.661}\)	&\
\(.762_{\pm.019}\)	&\(.858_{\pm.013}\)	&\(1.472_{\pm.103}\)	&\(.614_{\pm.037}\)	&\(\underline{3.623_{\pm.887}}\)	\\
Skea-topo	&\(.620_{\pm.095}\)	&\(.734_{\pm.085}\)	&\(2.236_{\pm.824}\)	&\(.463_{\pm.159}\)	&\(11.653_{\pm5.620}\)	&\
\(.784_{\pm.016}\)	&\(.873_{\pm.010}\)	&\(1.299_{\pm.086}\)	&\(.645_{\pm.022}\)	&\(3.363_{\pm.609}\)	\\
RMI	&\(\underline{.766_{\pm.008}}\)	&\(.852_{\pm.006}\)	&\(1.824_{\pm.263}\)	&\(.499_{\pm.065}\)	&\(11.632_{\pm3.918}\)	&\
\(.839_{\pm.013}\)	&\(.908_{\pm.008}\)	&\(.819_{\pm.071}\)	&\(.727_{\pm.032}\)	&\(2.587_{\pm.435}\)	\\
clDice	&\(.645_{\pm.051}\)	&\(.755_{\pm.045}\)	&\(2.162_{\pm.467}\)	&\(.476_{\pm.081}\)	&\(15.461_{\pm4.999}\)	&\
\(.800_{\pm.013}\)	&\(.884_{\pm.009}\)	&\(1.004_{\pm.113}\)	&\(.684_{\pm.026}\)	&\(3.146_{\pm.286}\)	\\
Sensitive	&\(.734_{\pm.022}\)	&\(.827_{\pm.018}\)	&\(2.933_{\pm.982}\)	&\(.305_{\pm.159}\)	&\(11.292_{\pm3.828}\)	&\
\(.824_{\pm.018}\)	&\(.899_{\pm.011}\)	&\(.880_{\pm.108}\)	&\(.702_{\pm.041}\)	&\(2.757_{\pm.398}\)	\\

\hline

CWMI-Real	&\(.765_{\pm.010}\)	&\(\underline{.853_{\pm.008}}\)	&\(\underline{1.231_{\pm.183}}\)	&\(\underline{.650_{\pm.050}}\)	&\(\underline{1.836_{\pm4.994}}\)	&\
\(\underline{.846_{\pm.012}}\)	&\(\underline{.913_{\pm.008}}\)	&\(\underline{.766_{\pm.067}}^{**}\)	&\(\underline{.736_{\pm.021}}^{*}\)	&\(2.504_{\pm.528}\)	\\
CWMI	&\(\mathbf{.767_{\pm.010}}\)	&\(\mathbf{.854_{\pm.008}}\)	&\(\mathbf{1.148_{\pm.166}}^{*}\)	&\(\mathbf{.667_{\pm.047}}^{*}\)	&\(\mathbf{1.326_{\pm4.109}}\)	&\
\(\mathbf{.839_{\pm.015}}\)	&\(\mathbf{.908_{\pm.010}}\)	&\(\mathbf{.762_{\pm.084}}^{***}\)	&\(\mathbf{.728_{\pm.035}}^{**}\)	&\(\mathbf{2.747_{\pm.680}}\)	\\

         \Xhline{1pt}
    \end{tabular}
    \caption{Quantitative results of different loss functions across the four datasets and two neural network models. The \textbf{bold} numbers indicate the best performance for each metric, while the \underline{underlined} numbers denote the second-best performance. A hyphen ("-") indicates cases where the model did not converge. For the loss metrics CWMI and CWMI-real, the statistical significance is marked with asterisks: *\(p<.05\), **\(p<.01\), and ***\(p<.001\), where p is the maximum p-value of the student's t tests against all the baseline loss functions.}
    \label{tab:1}
\end{table*}

\vspace{-10pt}
\section{Experiment}

\subsection{Experimental Setup}
\paragraph{Base Models}
To evaluate the effectiveness of the CWMI loss, we employed U-Net \cite{ronneberger2015u} and Attention U-Net \cite{oktay2018attention} as baseline architectures. This selection enables an assessment of CWMI’s generalization ability across both fully convolutional and attention-enhanced models. Additionally, to examine CWMI's compatibility with recently proposed Mamba-based architectures \cite{yue2024medmamba}, we incorporated the Vision Mamba U-Net (VMUNet) \cite{ruan2024vm} into our experiments.

\vspace{-10pt}
\paragraph{Datasets}
We tested CWMI on three public segmentation datasets, all characterized by class and instance imbalance: (1) SNEMI3D, a neurite segmentation dataset containing 100 \(1024 \times 1024\) grayscale images from electron microscopy slices \cite{arganda2013snemi3d}; (2) GlaS, a gland segmentation dataset with 165 RGB images of varying sizes from histological images of colorectal cancer samples \cite{sirinukunwattana2017gland}; (3) DRIVE, a retinal vessel segmentation dataset comprising 40 \(584 \times 565\) RGB images from fundus photographs \cite{staal2004ridge}; and (4) the Massachusetts Roads dataset (MASS ROAD), a road segmentation dataset with 1171 \(1500\times 1500\) RGB images from aerial imagery \cite{MnihThesis}. We choose a subset of 120 images (ignoring images without a network of roads). For all datasets, at least two repeated three-fold cross-validations were used to ensure robust evaluation.

\vspace{-10pt}
\paragraph{Baselines and Implementation Details}
We compared CWMI against 11 state-of-the-art (SOTA) loss functions, including pixel-wise loss functions (e.g., cross entropy, BCE \cite{long2015fully}, Focal loss \cite{ross2017focal}), Sensitive loss \cite{tang2025increase}, region-based loss functions (e.g., Dice loss \cite{milletari2016v}, Tversky loss \cite{salehi2017tversky}, Jaccard loss \cite{rahman2016optimizing}), and structural/topological loss functions (e.g., WCE \cite{ronneberger2015u}, ABW loss \cite{liu2022boundary}, Skea-topo loss \cite{liu2024enhancing}, RMI loss \cite{zhao2019region}, clDice loss \cite{shit2021cldice}). Hyperparameters for each baseline were tuned via grid search, with Tversky loss (\(\alpha = .5, \beta = .5\)) and Focal loss (\(\gamma = 2.5\)) as examples. 

We utilized steerable pyramids with four decomposition levels and four orientations in all experiments. \(N=4\) and \(K=4\) of the steerable pyramid decomposition and a regularization parameter of \(\lambda=0.1\) were determined by the ablation experiments, and applied in all CWMI experiments. For the implementation of Equation \ref{eq:10}, although PyTorch supports complex matrix calculations, our experiments indicated that its efficiency remains suboptimal. Consequently, we computed using their real representations, which are mathematically equivalent \cite{golub2013matrix}. To assess the significance of the complex steerable pyramid, we compared CWMI with a real-number-only variant (CWMI-Real), implemented according to Equations \ref{G real} and \ref{eq:9}. Adam optimizer with a StepLR scheduler (initial learning rate \(1 \times 10^{-4}\), decay rate 0.8, step size 10) was used. U-Net and Attention U-Net models were trained for 50 epochs; VMUNet models were trained for 100 epochs due to their slower convergence. All models were trained with a batch size of 10. Early stopping based on mIoU was employed to select the best model. Training was conducted on an NVIDIA A100 GPU using the Google Colab runtime.

\vspace{-10pt}
\paragraph{Data Augmentation and Evaluation Metrics}
Random flips and rotations were applied to all datasets to improve generalization. For SNEMI3D and MASS ROAD, images were randomly cropped to \(512 \times 512\), while for GlaS, images were cropped to \(448 \times 576\) to standardize input sizes. No cropping was performed for DRIVE due to its uniform image dimensions.

Performance was evaluated using five metrics: mIoU and mDice for regional precision, variation of information (VI) \cite{nunez2013machine} and adjusted Rand index (ARI) \cite{vinh2009information} for clustering precision, and Hausdorff distance (HD) for boundary and topological accuracy. These metrics provide a comprehensive assessment of both regional overlap and structural fidelity.

\begin{table}[h]
\scriptsize
    \centering
    \setlength{\tabcolsep}{1pt}
    \begin{tabular}{lccccc}
        \Xhline{1pt}
        \multicolumn{6}{c}{\textbf{SNEMI3D VMUNet}}\\
Methods  & 
mIoU$\uparrow$ & mDice$\uparrow$ & VI$\downarrow$ & ARI$\uparrow$ & HD$\downarrow$ \\
        \hline
CE	&\(.723_{\pm.026}\)	&\(.829_{\pm.020}\)	&\(2.577_{\pm1.211}\)	&\(.414_{\pm.173}\)	&\(1.709_{\pm.235}\)	\\
BCE	&\(.739_{\pm.004}\)	&\(.843_{\pm.002}\)	&\(1.499_{\pm.057}\)	&\(\underline{.594_{\pm.010}}\)	&\(1.274_{\pm.231}\)	\\
Dice	&\(.725_{\pm.029}\)	&\(.831_{\pm.021}\)	&\(2.735_{\pm1.213}\)	&\(.402_{\pm.165}\)	&\(1.767_{\pm.720}\)	\\
Focal	&\(.725_{\pm.010}\)	&\(.833_{\pm.007}\)	&\(1.677_{\pm.141}\)	&\(.563_{\pm.025}\)	&\(1.699_{\pm.372}\)	\\
Jaccard	&\(.725_{\pm.019}\)	&\(.831_{\pm.014}\)	&\(1.953_{\pm.541}\)	&\(.499_{\pm.075}\)	&\(1.716_{\pm.566}\)	\\
Tversky	&\(.737_{\pm.016}\)	&\(.840_{\pm.012}\)	&\(1.619_{\pm.324}\)	&\(.555_{\pm.045}\)	&\(1.560_{\pm.408}\)	\\
WCE	&\(.714_{\pm.005}\)	&\(.826_{\pm.004}\)	&\(1.743_{\pm.026}\)	&\(.552_{\pm.006}\)	&\(1.432_{\pm.169}\)	\\
ABW	&\(.678_{\pm.003}\)	&\(.800_{\pm.003}\)	&\(2.076_{\pm.030}\)	&\(.490_{\pm.005}\)	&\(1.572_{\pm.021}\)	\\
RMI	&\(\underline{.762_{\pm.010}}\)	&\(\underline{.857_{\pm.007}}\)	&\(\underline{1.419_{\pm.183}}\)	&\(.588_{\pm.030}\)	&\(\underline{1.071_{\pm.306}}\)	\\
clDice	&\(.714_{\pm.002}\)	&\(.824_{\pm.001}\)	&\(1.754_{\pm.062}\)	&\(.542_{\pm.003}\)	&\(1.990_{\pm.262}\)	\\
sensitive	&\(-\)	&\(-\)	&\(-\)	&\(-\)	&\(-\) \\

\hline

CWMI	&\(\mathbf{.783_{\pm.004}}^{*}\)	&\(\mathbf{.872_{\pm.003}}^{*}\)	&\(\mathbf{.982_{\pm.106}}^{*}\)	&\(\mathbf{.660_{\pm.022}}^{*}\)	&\(\mathbf{.629_{\pm.073}}\)	\\

\Xhline{1pt}
\end{tabular}
    \caption{Quantitative results of different loss functions on SNEMI3D dataset and VMUNet model. The \textbf{bold} numbers indicate the best performance for each metric, while the \underline{underlined} numbers denote the second-best performance. A hyphen ("-") indicates cases where the model did not converge. For the loss metrics CWMI, the statistical significance is marked with asterisks: *\(p<.05\), where p is the maximum p-value of the student's t tests against all the baseline loss functions.}
    \label{tab:2}
\end{table}

\begin{figure}[!h]
    \centering
    \includegraphics[width=\linewidth]{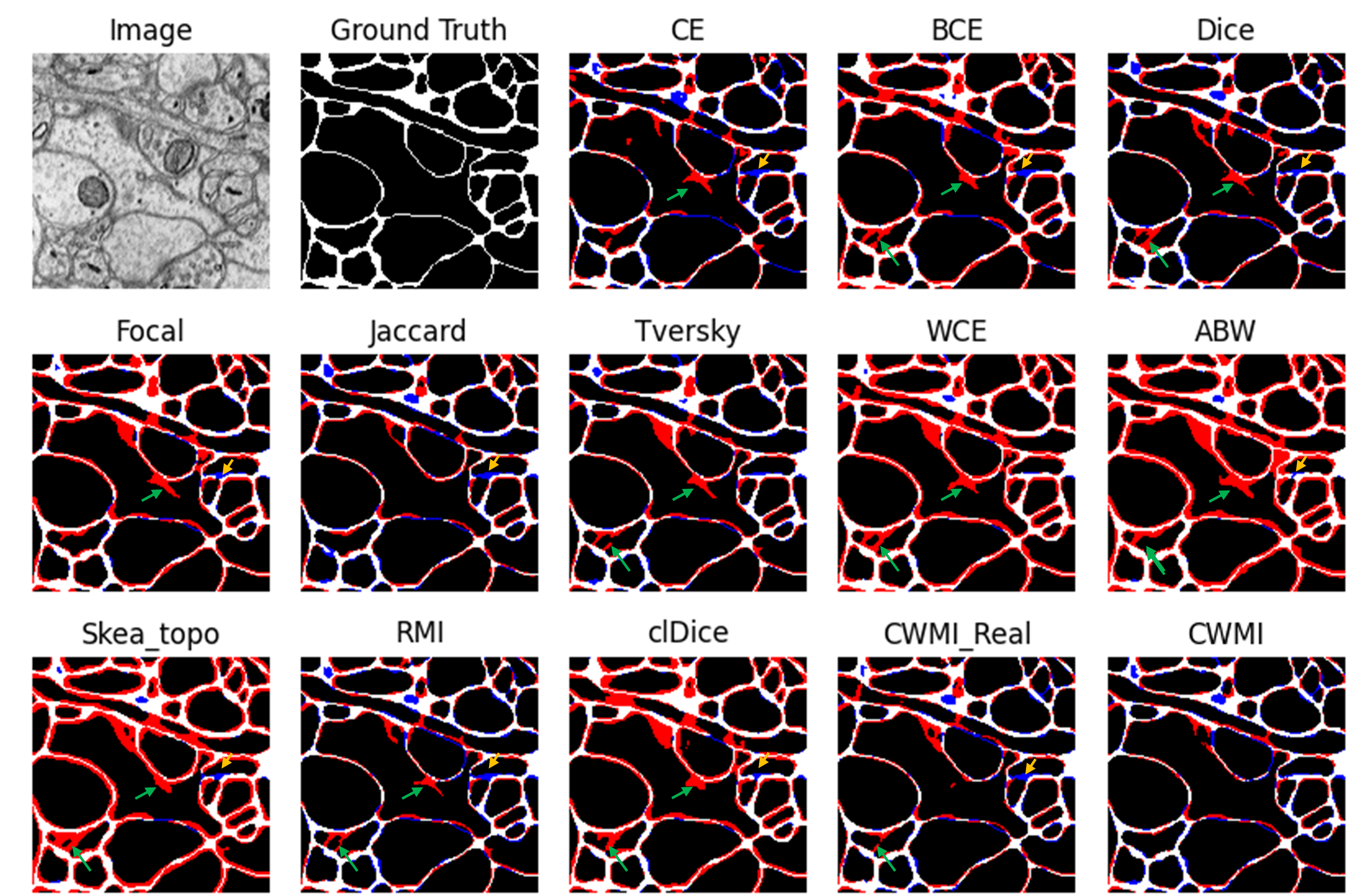}
    \vspace{-20pt}
    \caption{Qualitative results of different loss functions on the SNEMI3D dataset. \textcolor{red}{Red: false positive regions;} \textcolor{blue}{Blue: false negative regions.} \textcolor{ForestGreen}{Green arrow: challenging false positive} and \textcolor{orange}{Orange arrow: challenging false negative} that are successfully addressed by CWMI.}
    \label{fig:3}
\end{figure}
\begin{figure}[!h]
    \centering
    \includegraphics[width=\linewidth]{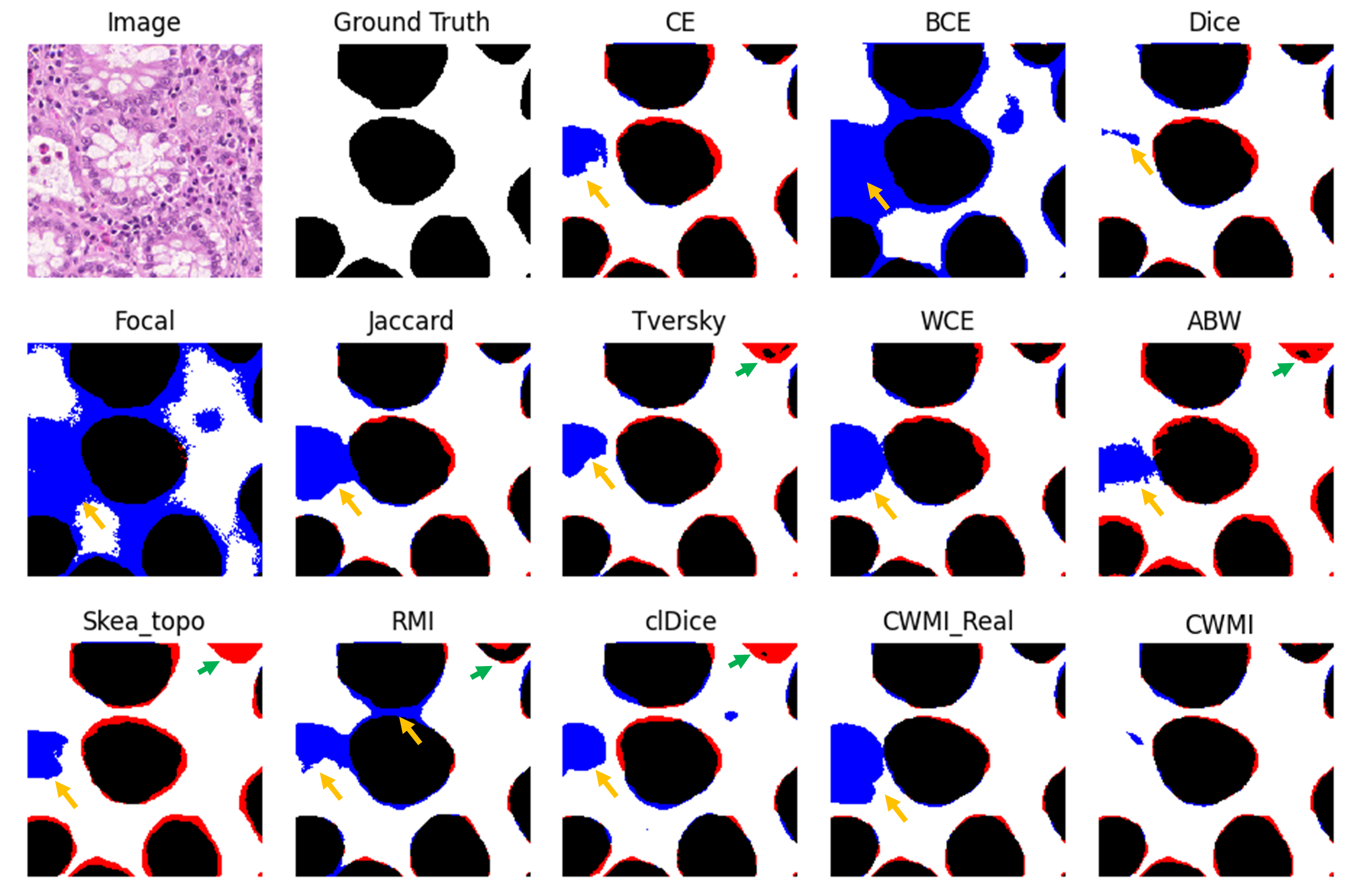}
    \vspace{-20pt}
    \caption{Qualitative results of different loss functions on the GlaS dataset. }
    \label{fig:4}
\end{figure}

\begin{figure}[!h]
    \centering
    \includegraphics[width=\linewidth]{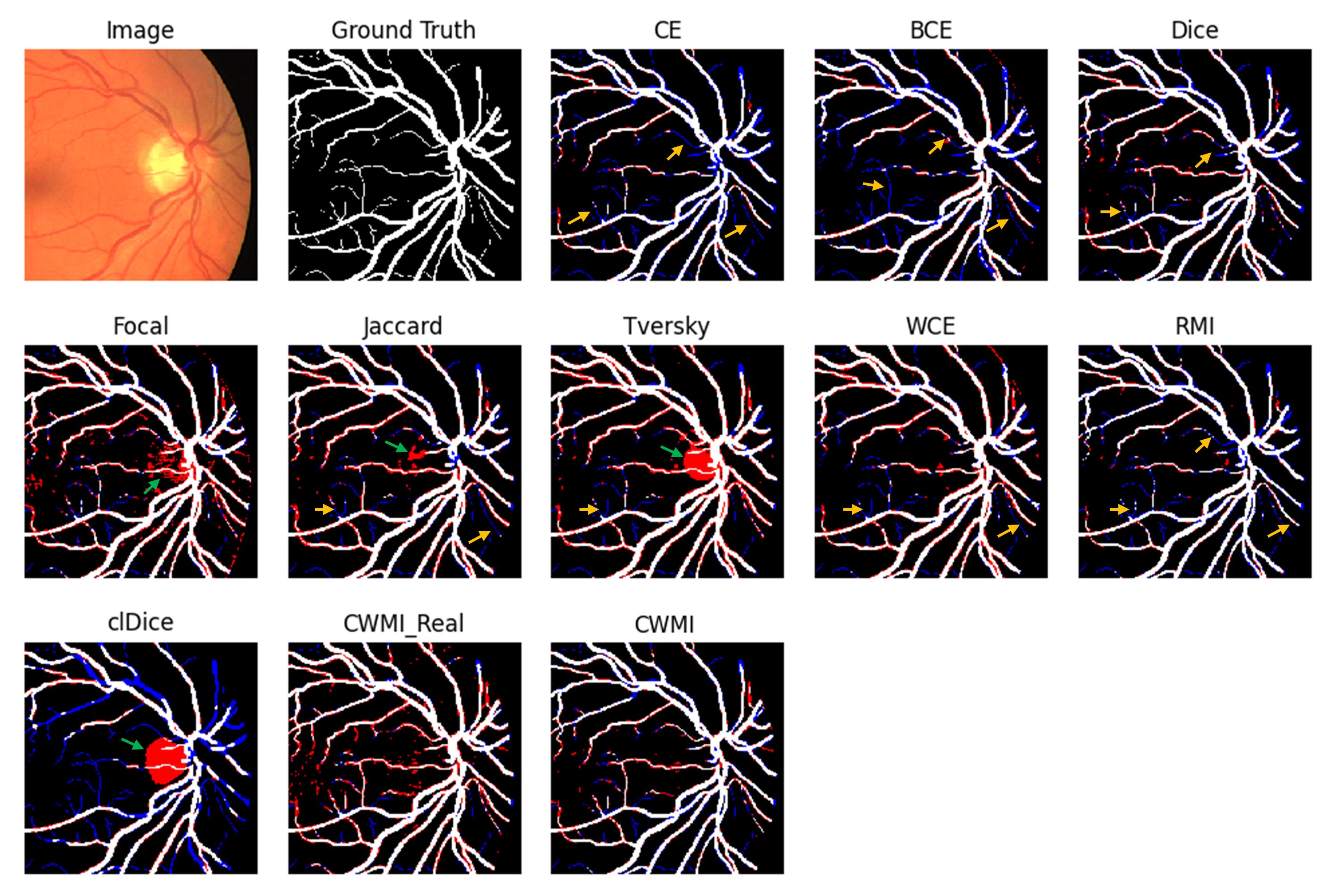}
    \vspace{-20pt}
    \caption{Qualitative results of different loss functions on the DRIVE dataset.}
    \label{fig:5}
\end{figure}

\vspace{-10pt}
\subsection{Quantitative and qualitative results}
\begin{figure}[!h]
    \centering
    \includegraphics[width=\linewidth]{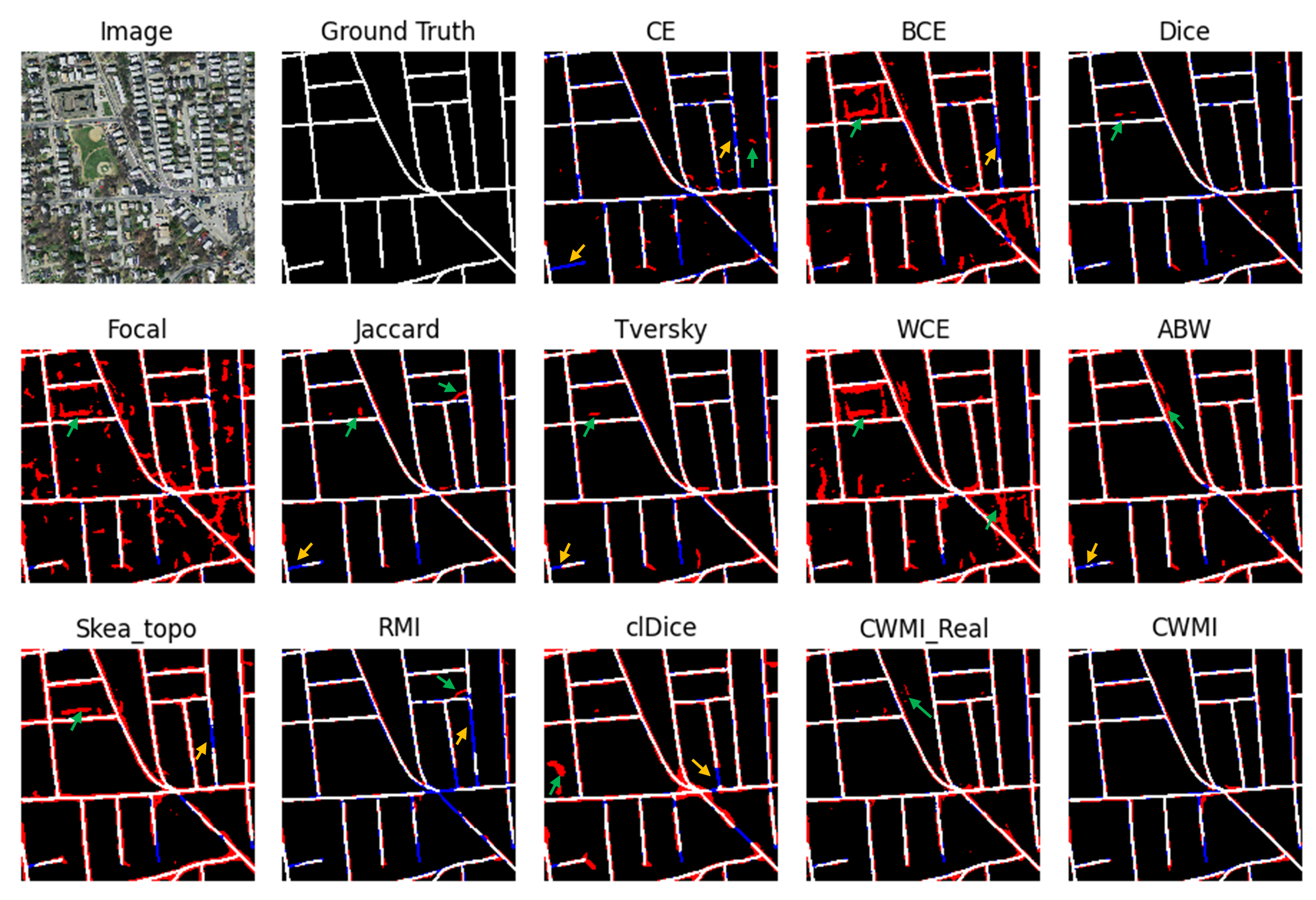}
    \vspace{-20pt}
    \caption{Qualitative results of different loss functions on the MASS ROAD dataset.}
    \label{fig:6}
\end{figure}

As shown in Table \ref{tab:1}, the proposed CWMI loss outperforms other loss functions across the majority of evaluation metrics for all datasets using both U-Net and Attention U-Net architectures. Statistically, CWMI achieves significantly superior performance on all metrics for the SNEMI3D and DRIVE datasets, as well as on clustering-based metrics (VI and ARI) for the GlaS and MASS ROAD datasets. Moreover, CWMI consistently outperforms its real-valued variant, underscoring the importance of incorporating phase information within the steerable pyramid decomposition. For the Mamba-based architecture, VMUNet, CWMI also significantly outperforms baseline loss functions on the SNEMI3D dataset (Table \ref{tab:2}). Qualitative results from SNEMI3D (Figure \ref{fig:3}), GlaS (Figure \ref{fig:4}), DRIVE (Figure \ref{fig:5}), and MASS ROAD (Figure \ref{fig:6}) further illustrate CWMI’s effectiveness in mitigating challenging false positive and false negative segmentation errors that persist with other state-of-the-art loss functions.

\vspace{-15pt}
\subsection{Ablation studies}
\vspace{-5pt}
In this section, all ablation experiments were performed on the SNEMI3D dataset with U-Net model.

\vspace{-10pt}
\paragraph{Mutual Information (MI) vs \(\mathbf{L_1}\), \(\mathbf{L_2}\) Distance and Structural Similarity (SSIM)} As previously discussed, various metrics based on wavelet transforms have been developed for evaluating image similarity and guiding loss functions. To assess the advantages of mutual information (MI) over commonly used metrics such as \(L_1\), \(L_2\), and SSIM, we conducted a comparative analysis. The experimental results, summarized in Table \ref{tab:3}, demonstrate that MI consistently outperforms \(L_1\), \(L_2\) distance, and SSIM across multiple evaluation metrics. Unlike \(L_1\) and \(L_2\), which focus on pixel-wise intensity differences, and SSIM, which emphasizes structural similarity, MI captures joint statistical dependencies between features in each direction. This ability provides a more robust representation of structural differences between predictions and labels, particularly in complex segmentation tasks.

\begin{table}[h]
\scriptsize
    \centering
    \setlength{\tabcolsep}{1pt}
    \begin{tabular}{lccccc}
        \Xhline{1pt}
Methods  & 
mIoU$\uparrow$ & mDice$\uparrow$ & VI$\downarrow$ & ARI$\uparrow$ & HD$\downarrow$ \\
        \hline
L1 & \(.773_{\pm.002}\) & \(.866_{\pm.001}\) & \(1.22_{\pm.05}\) & \(.632_{\pm.007}\) & \(.70_{\pm.08}\) \\
L2 & \(.774_{\pm.003}\) & \(.867_{\pm.002}\) & \(1.23_{\pm.06}\) & \(.633_{\pm.011}\) & \(.76_{\pm.13}\) \\
SSIM & \(.468_{\pm.032}\) & \(.613_{\pm.020}\) & \(5.96_{\pm.96}\) & \(.081_{\pm.030}\) & \(9.31_{\pm1.64}\) \\
CWMI & \(\mathbf{.779_{\pm.004}}\) & \(\mathbf{.870_{\pm.003}}\) & \(\mathbf{1.16_{\pm.09}}\) & \(\mathbf{.640_{\pm.005}}\) & \(\mathbf{.69_{\pm.04}}\) \\
\Xhline{1pt}
\end{tabular}
    \caption{Quantitative results comparing mutual information (MI), \(L_1\), \(L_2\) distance, and structural similarity (SSIM) based on U-Net with SNEMI3D. The \textbf{bold} numbers indicate the best performance for each metric.}
    \label{tab:3}
\end{table}

\vspace{-10pt}
\paragraph{Contribution of different components of complex steerable pyramid} To better understand the role of individual components in the CWMI loss, we conducted an ablation experiment evaluating the performance of variants based on the mutual information loss of magnitude-only, phase-only, and real-only representations, as shown in Table \ref{tab:4}. The CWMI-Phase variant consistently underperformed across all metrics, suggesting that phase information alone is insufficient for reliable segmentation. The CWMI-Mag variant, by contrast, achieved the best performance on clustering metrics (VI and ARI), indicating its strong capacity to capture regional structures. The full CWMI, combining both magnitude and phase via the complex representation, yielded the best results on pixel-wise accuracy metrics (mIoU and mDice) and topological metric (HD), while ranking second on VI and ARI, demonstrating a well-rounded performance across different dimensions of segmentation quality. The CWMI-Real variant performed worse than both CWMI and CWMI-Mag on most metrics, although it outperformed CWMI-Phase, further supporting the necessity of incorporating both magnitude and phase components. These results highlight the critical role of magnitude in structural segmentation, while also demonstrating that the full complex formulation of CWMI achieves superior generalization across diverse evaluation criteria. 

\begin{table}[h]
\scriptsize
    \centering
    \setlength{\tabcolsep}{1pt}
    \begin{tabular}{lccccc}
        \Xhline{1pt}
Methods  & 
mIoU$\uparrow$ & mDice$\uparrow$ & VI$\downarrow$ & ARI$\uparrow$ & HD$\downarrow$ \\
        \hline
CWMI-Real	&\(\underline{.776_{\pm.004}}\)	&\(\underline{.868_{\pm.003}}\)	&\(1.211_{\pm.048}\)	&\(.635_{\pm.011}\)	&\(.774_{\pm.065}\)	\\
CWMI-Mag	&\(.775_{\pm.004}\)	&\(.867_{\pm.003}\)	&\(\mathbf{.908_{\pm.051}}\)	&\(\mathbf{.660_{\pm.009}}\)	&\(\underline{.735_{\pm.065}}\)	\\
CWMI-Phase	&\(.694_{\pm.008}\)	&\(.810_{\pm.006}\)	&\(2.487_{\pm.534}\)	&\(.458_{\pm.061}\)	&\(2.000_{\pm.258}\)	\\
CWMI	&\(\mathbf{.778_{\pm.004}}\)	&\(\mathbf{.869_{\pm.003}}\)	&\(\underline{1.164_{\pm.079}}\)	&\(\underline{.638_{\pm.016}}\)	&\(\mathbf{.720_{\pm.066}}\)	\\

\Xhline{1pt}
\end{tabular}
    \caption{Quantitative comparison of mutual information loss across different components of the complex steerable pyramid decomposition. CWMI-Real: real component; CWMI-Mag: magnitude component; CWMI-Phase: phase component; CWMI: full complex representation (combining real and imaginary components). Results are based on a U-Net model trained on the SNEMI3D dataset. \textbf{Bold} values indicate the best performance, and \underline{underlined} values denote the second-best for each evaluation metric.}
    \label{tab:4}
\end{table}

\vspace{-10pt}
\paragraph{Impact of Regularization Parameter \(\lambda\)}
To evaluate the sensitivity of CWMI to the regularization weight \(\lambda\), we experimented with \(\lambda=0.1,0.5, 0.9\). As shown in Table~\ref{tab:5}, performance across all evaluation metrics remained largely stable, with only marginal variations. This robustness suggests that CWMI provides complementary information to the cross-entropy term and is not overly sensitive to the regularization strength. Additionally, we found the inclusion of the cross-entropy term to be crucial for directional learning, as mutual information is inherently symmetric and cannot distinguish between correct and inverted label assignments.

\vspace{-10pt}
\paragraph{Impact of Decomposition Level \(N\) and Number of Orientations \(K\)} 
Theoretically, the steerable pyramid can decompose an image into very high levels with an infinite number of orientations, provided the input image is sufficiently large. However, does deeper decomposition or a higher number of orientations improve feature extraction and loss computation? To address this, we analyzed the impact of \(N\) (decomposition level) and \(K\) (number of orientations) on the performance of CWMI, as shown in Table \ref{tab:3}. Interestingly, both \(N\) and \(K\) achieved optimal performance at relatively low values, suggesting that the critical information for segmentation is concentrated in relatively high-frequency regions. This observation supports the idea that emphasizing high-frequency subbands refines small instances and narrow boundaries, while lower-frequency components from deeper decompositions contribute less informative features compared to higher-frequency ones.

Knowing that the first four layers of decomposition are crucial, we further performed a layer ablation experiment (Table \ref{tab:3}, Layer Ablation), where only selected layers were used to compute mutual information while discarding the remaining layers. Our results show that the third layer outperforms other layers in region-based (mIoU, mDice) and cluster-based metrics (VI, ARI), while the fourth layer achieves the best performance in topological metrics (HD). These findings suggest that features extracted from multiple layers are essential for achieving high performance across all evaluation metrics, highlighting the importance of incorporating information from both mid- and high-frequency subbands in the segmentation task.

\begin{table}[h]
\scriptsize
    \centering
    \setlength{\tabcolsep}{1pt}
    \begin{tabular}{lccccc}
        \Xhline{1pt}
        \multicolumn{6}{c}{\(\mathbf{\lambda}\) \textbf{ablation}}\\ 
        \hline
        ~  & 
mIoU$\uparrow$ & mDice$\uparrow$ & VI$\downarrow$ & ARI$\uparrow$ & HD$\downarrow$ \\
        \hline
    \(\lambda\)=0.1	&\(\mathbf{.778_{\pm.004}}\)	&\(\mathbf{.869_{\pm.003}}\)	&\(\mathbf{1.188_{\pm.075}}\)	&\(.636_{\pm.008}\)	&\(\mathbf{.703_{\pm.034}}\)	\\
\(\lambda\)=0.5	&\(.778_{\pm.006}\)	&\(.869_{\pm.004}\)	&\(1.176_{\pm.076}\)	&\(\mathbf{.639_{\pm.014}}\)	&\(.729_{\pm.078}\)	\\
\(\lambda\)=0.9	&\(.777_{\pm.003}\)	&\(.869_{\pm.002}\)	&\(1.219_{\pm.073}\)	&\(.636_{\pm.013}\)	&\(.740_{\pm.061}\)	\\
\Xhline{1pt}

        \multicolumn{6}{c}{\(\mathbf{K}\)  \textbf{ablation}}\\
        \hline
~  & 
mIoU$\uparrow$ & mDice$\uparrow$ & VI$\downarrow$ & ARI$\uparrow$ & HD$\downarrow$ \\
\hline
K=2 & \(.777_{\pm.006}\) & \(.868_{\pm.004}\) & \(1.18_{\pm.10}\) & \(.637_{\pm.028}\) & \(.79_{\pm.02}\) \\
K=4 & \(\mathbf{.779_{\pm.004}}\) & \(\mathbf{.870_{\pm.003}}\) & \(\mathbf{1.16_{\pm.09}}\) & \(\mathbf{.640_{\pm.005}}\) & \(\mathbf{.69_{\pm.04}}\) \\
K=8 & \(.776_{\pm.003}\) & \(.868_{\pm.002}\) & \(1.20_{\pm.06}\) & \(.635_{\pm.009}\) & \(.76_{\pm.08}\) \\
K=12 & \(.772_{\pm.005}\) & \(.865_{\pm.003}\) & \(1.33_{\pm.05}\) & \(.621_{\pm.004}\) & \(.81_{\pm.08}\) \\
\Xhline{1pt}
        \multicolumn{6}{c}{\(\mathbf{N}\) \textbf{ablation}}\\
        \hline
~  & 
mIoU$\uparrow$ & mDice$\uparrow$ & VI$\downarrow$ & ARI$\uparrow$ & HD$\downarrow$ \\
\hline
N=2 & \(.760_{\pm.002}\) & \(.857_{\pm.001}\) & \(1.30_{\pm.04}\) & \(.618_{\pm.009}\) & \(1.20_{\pm.12}\) \\
N=4 & \(\mathbf{.777_{\pm.002}}\) & \(\mathbf{.869_{\pm.001}}\) & \(\mathbf{1.21_{\pm.10}}\) & \(\mathbf{.634_{\pm.017}}\) & \(\mathbf{.79_{\pm.190}}\) \\
N=6 & \(.768_{\pm.003}\) & \(.862_{\pm.002}\) & \(1.29_{\pm.05}\) & \(.618_{\pm.004}\) & \(.88_{\pm.18}\) \\
\Xhline{1pt}
        \multicolumn{6}{c}{\textbf{Layer ablation}}\\
        \hline
~  & 
mIoU$\uparrow$ & mDice$\uparrow$ & VI$\downarrow$ & ARI$\uparrow$ & HD$\downarrow$ \\
\hline
1st layer & \(.738_{\pm.008}\) & \(.843_{\pm.006}\) & \(1.61_{\pm.07}\) & \(.576_{\pm.015}\) & \(1.38_{\pm.14}\) \\
2nd layer & \(.762_{\pm.008}\) & \(.858_{\pm.006}\) & \(1.25_{\pm.08}\) & \(.624_{\pm.016}\) & \(1.12_{\pm.19}\) \\
3rd layer & \(\mathbf{.774_{\pm.003}}\) & \(\mathbf{.867_{\pm.002}}\) & \(\mathbf{1.23_{\pm.10}}\) & \(\mathbf{.638_{\pm.020}}\) & \(.82_{\pm.12}\) \\
4th layer & \(.766_{\pm.004}\) & \(.861_{\pm.003}\) & \(1.39_{\pm.09}\) & \(.614_{\pm.012}\) & \(\mathbf{.76_{\pm.04}}\) \\

\Xhline{1pt}
\end{tabular}
    \caption{The impact of regularization parameter \(\lambda\), decomposition level N, and orientation number K on CWMI performance on U-Net with SNEMI3D. In the layer ablation test, only the selected layer is computed for mutual information and all other layers are discard. The \textbf{bold} numbers indicate the best performance for each metric.}
    \label{tab:5}
\end{table}

\vspace{-10pt}
\paragraph{Computational complexity analysis} 
For an input image of size \( H \times W \) with \( K \) orientation decompositions, the computational complexity of the CWMI loss function is analyzed as follows.

First, the forward Fourier transform has a time complexity of \(O(HW \log(HW))\). For the first layer of decomposition, the operations include:
Band-pass filtering: \( O(HW K) \);
Inverse Fourier transform: \( O(HW \log(HW) K) \);
and mutual information computation: \( O(HW K^2) \).
Summing these terms, the total complexity for the first decomposition layer is:
\(
O(HW (K \log(HW) + K^2))
\)
For subsequent decomposition layers, the image size reduces by a factor of four at each step, meaning the second layer processes an image of size \( HW / 4 \), and the third layer processes \( HW / 16 \), and so on. Thus, the total computational complexity follows a geometric series with a superior bound 
\(
O\left(\frac{4}{3} HW (K \log(HW) + K^2) \right)
\)

From our ablation experiments, the optimal number of orientations is \( K = 4 \), which is relatively small. Hence the complexity of CWMI is linear to 
\(
O(HW \log(HW))
\), which remains scalable for high-resolution images. Compared to topology-aware losses such as RMI, clDice, or Hausdorff Distance Loss, CWMI achieves better structural and boundary preservation with lower computational overhead, as shown in Table \ref{tab:4}.

\begin{table}[h]
\scriptsize
    \centering
    \setlength{\tabcolsep}{1pt}
    \begin{tabular}{lcc}
        \Xhline{1pt}
        ~ & Epoch Time (s) & \(\Delta t\) to CE (s) \\
        \hline
        CE & 2.04 & .00 \\
BCE & 2.02 & -0.02 \\
Dice & 2.01 & -0.03 \\
Focal & 2.01 & -0.03 \\
Jaccard & 2.02 & -0.02 \\
Tversky & 2.01 & -0.03 \\
WCE & 2.15 & 0.12 \\
ABW & 2.33 & 0.30 \\
Skea-topo & 2.72 & 0.69 \\
RMI & 2.28 & 0.25 \\
clDice & 2.37 & 0.34 \\
\hline
CWMI-Real & 2.18 & 0.15 \\
CWMI & 2.27 & 0.23 \\
        \Xhline{1pt}
    \end{tabular}
    \caption{Training time per epoch and relative change to CE baseline for various loss functions on U-Net model with SNEMI3D dataset.}
    \label{tab:6}
\end{table}

\vspace{-10pt}
\section{Conclusion}
\vspace{-5pt}

In this study, we introduced Complex Wavelet Mutual Information (CWMI) loss, a novel loss function for semantic segmentation that leverages the multiscale, multi-orientation decomposition capabilities of the complex steerable pyramid. By integrating mutual information across wavelet subbands, CWMI effectively captures high-dimensional dependencies and local structural features, including critical phase information, which are essential for accurate segmentation. Extensive experiments on four challenging datasets demonstrate that CWMI consistently outperforms state-of-the-art loss functions across most evaluation metrics, particularly in segmenting small instances and narrow boundaries, while introducing minimal computational overhead. These results highlight CWMI as a robust and versatile loss function that effectively addresses key challenges in segmentation, such as class and instance imbalance, boundary precision, and topological consistency.

Beyond semantic segmentation, the core principles of CWMI—multiscale feature extraction and structural consistency—suggest its potential applicability to a broader range of computer vision and machine learning tasks, such as image-to-image translation and super-resolution, which we leave for future exploration.

Although extensively studied on 2D images, extending it to multi-class and 3D segmentation is theoretically feasible but necessitates further validation in future research. 

\section*{Acknowledgements}

This study was conducted without external funding. I would like to thank the reviewers for their constructive feedback, which significantly improved the quality of this work. I am especially grateful to my wife, Di Duan, for her unwavering support of my passion for machine learning. I thank Xinzi He for valuable discussions that contributed to the revision of this study.

\section*{Impact Statement}

The proposed Complex Wavelet Mutual Information (CWMI) loss introduces a novel approach to structural-aware learning in deep neural networks. By leveraging multiscale decomposition through the complex steerable pyramid and mutual information across frequency subbands, CWMI enables improved segmentation performance, particularly for small-scale structures and thin boundaries. Our empirical results demonstrate significant improvements in both pixel-wise and topological accuracy across multiple datasets. Beyond segmentation, CWMI has the potential to generalize to a wide range of real-world applications, including medical imaging, autonomous driving, satellite-based environmental monitoring, and industrial defect detection. 

Overall, CWMI provides a \textbf{computationally efficient}, \textbf{structure-aware} loss function that enhances segmentation performance while maintaining practical scalability. By extending its application beyond segmentation, we aim to contribute to the broader field of generative modeling, object detection, and self-supervised learning in deep neural networks.

\bibliography{main_bib}

\begin{thebibliography}{48}
\providecommand{\natexlab}[1]{#1}
\providecommand{\url}[1]{\texttt{#1}}
\expandafter\ifx\csname urlstyle\endcsname\relax
  \providecommand{\doi}[1]{doi: #1}\else
  \providecommand{\doi}{doi: \begingroup \urlstyle{rm}\Url}\fi

\bibitem[Arganda-Carreras et~al.(2013)Arganda-Carreras, Seung, Vishwanathan, and Berger]{arganda2013snemi3d}
Arganda-Carreras, I., Seung, H.~S., Vishwanathan, A., and Berger, D.~R.
\newblock Snemi3d: 3d segmentation of neurites in em images (isbi 2013).
\newblock \url{https://snemi3d.grand-challenge.org/}, 2013.
\newblock Accessed: 2024-12-20.

\bibitem[Azad et~al.(2023)Azad, Heidary, Yilmaz, H{\"u}ttemann, Karimijafarbigloo, Wu, Schmeink, and Merhof]{azad2023loss}
Azad, R., Heidary, M., Yilmaz, K., H{\"u}ttemann, M., Karimijafarbigloo, S., Wu, Y., Schmeink, A., and Merhof, D.
\newblock Loss functions in the era of semantic segmentation: A survey and outlook.
\newblock \emph{arXiv preprint arXiv:2312.05391}, 2023.

\bibitem[Belghazi et~al.(2018)Belghazi, Baratin, Rajeshwar, Ozair, Bengio, Courville, and Hjelm]{belghazi2018mutual}
Belghazi, M.~I., Baratin, A., Rajeshwar, S., Ozair, S., Bengio, Y., Courville, A., and Hjelm, D.
\newblock Mutual information neural estimation.
\newblock In \emph{International conference on machine learning}, pp.\  531--540. PMLR, 2018.

\bibitem[Canny(1986)]{canny1986computational}
Canny, J.
\newblock A computational approach to edge detection.
\newblock \emph{IEEE Transactions on pattern analysis and machine intelligence}, \penalty0 (6):\penalty0 679--698, 1986.

\bibitem[Chen et~al.(2021)Chen, Lu, Yu, Luo, Adeli, Wang, Lu, Yuille, and Zhou]{chen2021transunet}
Chen, J., Lu, Y., Yu, Q., Luo, X., Adeli, E., Wang, Y., Lu, L., Yuille, A.~L., and Zhou, Y.
\newblock Transunet: Transformers make strong encoders for medical image segmentation.
\newblock \emph{arXiv preprint arXiv:2102.04306}, 2021.

\bibitem[Diakogiannis et~al.(2020)Diakogiannis, Waldner, Caccetta, and Wu]{diakogiannis2020resunet}
Diakogiannis, F.~I., Waldner, F., Caccetta, P., and Wu, C.
\newblock Resunet-a: A deep learning framework for semantic segmentation of remotely sensed data.
\newblock \emph{ISPRS Journal of Photogrammetry and Remote Sensing}, 162:\penalty0 94--114, 2020.

\bibitem[Golub \& Van~Loan(2013)Golub and Van~Loan]{golub2013matrix}
Golub, G.~H. and Van~Loan, C.~F.
\newblock \emph{Matrix computations}.
\newblock JHU press, 2013.

\bibitem[Hjelm et~al.(2018)Hjelm, Fedorov, Lavoie-Marchildon, Grewal, Bachman, Trischler, and Bengio]{hjelm2018learning}
Hjelm, R.~D., Fedorov, A., Lavoie-Marchildon, S., Grewal, K., Bachman, P., Trischler, A., and Bengio, Y.
\newblock Learning deep representations by mutual information estimation and maximization.
\newblock \emph{arXiv preprint arXiv:1808.06670}, 2018.

\bibitem[Huang et~al.(2019)Huang, Huang, Yuan, and Kong]{huang2019fixed}
Huang, M., Huang, C., Yuan, J., and Kong, D.
\newblock Fixed-point deformable u-net for pancreas ct segmentation.
\newblock In \emph{Proceedings of the Third International Symposium on Image Computing and Digital Medicine}, pp.\  283--287, 2019.

\bibitem[Islam et~al.(2020)Islam, Vibashan, Jose, Wijethilake, Utkarsh, and Ren]{islam2020brain}
Islam, M., Vibashan, V., Jose, V. J.~M., Wijethilake, N., Utkarsh, U., and Ren, H.
\newblock Brain tumor segmentation and survival prediction using 3d attention unet.
\newblock In \emph{Brainlesion: Glioma, Multiple Sclerosis, Stroke and Traumatic Brain Injuries: 5th International Workshop, BrainLes 2019, Held in Conjunction with MICCAI 2019, Shenzhen, China, October 17, 2019, Revised Selected Papers, Part I 5}, pp.\  262--272. Springer, 2020.

\bibitem[Jiang et~al.(2024)Jiang, Li, Yi, Chen, and Wang]{jiang2024multi}
Jiang, W., Li, Y., Yi, Z., Chen, M., and Wang, J.
\newblock Multi-instance imbalance semantic segmentation by instance-dependent attention and adaptive hard instance mining.
\newblock \emph{Knowledge-Based Systems}, 304:\penalty0 112554, 2024.

\bibitem[Karimi \& Salcudean(2019)Karimi and Salcudean]{karimi2019reducing}
Karimi, D. and Salcudean, S.~E.
\newblock Reducing the hausdorff distance in medical image segmentation with convolutional neural networks.
\newblock \emph{IEEE Transactions on medical imaging}, 39\penalty0 (2):\penalty0 499--513, 2019.

\bibitem[Kervadec et~al.(2019)Kervadec, Bouchtiba, Desrosiers, Granger, Dolz, and Ayed]{kervadec2019boundary}
Kervadec, H., Bouchtiba, J., Desrosiers, C., Granger, E., Dolz, J., and Ayed, I.~B.
\newblock Boundary loss for highly unbalanced segmentation.
\newblock In \emph{International conference on medical imaging with deep learning}, pp.\  285--296. PMLR, 2019.

\bibitem[Kim \& Cho(2023)Kim and Cho]{kim2023whfl}
Kim, M.~W. and Cho, N.~I.
\newblock Whfl: Wavelet-domain high frequency loss for sketch-to-image translation.
\newblock In \emph{Proceedings of the IEEE/CVF Winter Conference on applications of computer vision}, pp.\  744--754, 2023.

\bibitem[Kofler et~al.(2023)Kofler, Shit, Ezhov, Fidon, Horvath, Al-Maskari, Li, Bhatia, Loehr, Piraud, et~al.]{kofler2023blob}
Kofler, F., Shit, S., Ezhov, I., Fidon, L., Horvath, I., Al-Maskari, R., Li, H.~B., Bhatia, H., Loehr, T., Piraud, M., et~al.
\newblock Blob loss: Instance imbalance aware loss functions for semantic segmentation.
\newblock In \emph{International Conference on Information Processing in Medical Imaging}, pp.\  755--767. Springer, 2023.

\bibitem[Korkmaz \& Tekalp(2024)Korkmaz and Tekalp]{korkmaz2024training}
Korkmaz, C. and Tekalp, A.~M.
\newblock Training transformer models by wavelet losses improves quantitative and visual performance in single image super-resolution.
\newblock \emph{arXiv preprint arXiv:2404.11273}, 2024.

\bibitem[Lin et~al.(2017)Lin, Doll{\'a}r, Girshick, He, Hariharan, and Belongie]{lin2017feature}
Lin, T.-Y., Doll{\'a}r, P., Girshick, R., He, K., Hariharan, B., and Belongie, S.
\newblock Feature pyramid networks for object detection.
\newblock In \emph{Proceedings of the IEEE conference on computer vision and pattern recognition}, pp.\  2117--2125, 2017.

\bibitem[Liu et~al.(2024)Liu, Ma, Ban, Xie, Wang, Xue, Ma, and Xu]{liu2024enhancing}
Liu, C., Ma, B., Ban, X., Xie, Y., Wang, H., Xue, W., Ma, J., and Xu, K.
\newblock Enhancing boundary segmentation for topological accuracy with skeleton-based methods.
\newblock \emph{arXiv preprint arXiv:2404.18539}, 2024.

\bibitem[Liu et~al.(2022)Liu, Chen, Liu, Ban, Ma, Wang, Xue, and Guo]{liu2022boundary}
Liu, W., Chen, J., Liu, C., Ban, X., Ma, B., Wang, H., Xue, W., and Guo, Y.
\newblock Boundary learning by using weighted propagation in convolution network.
\newblock \emph{Journal of Computational Science}, 62:\penalty0 101709, 2022.

\bibitem[Long et~al.(2015)Long, Shelhamer, and Darrell]{long2015fully}
Long, J., Shelhamer, E., and Darrell, T.
\newblock Fully convolutional networks for semantic segmentation.
\newblock In \emph{Proceedings of the IEEE conference on computer vision and pattern recognition}, pp.\  3431--3440, 2015.

\bibitem[Mallat(1989)]{mallat1989theory}
Mallat, S.~G.
\newblock A theory for multiresolution signal decomposition: the wavelet representation.
\newblock \emph{IEEE transactions on pattern analysis and machine intelligence}, 11\penalty0 (7):\penalty0 674--693, 1989.

\bibitem[Milletari et~al.(2016)Milletari, Navab, and Ahmadi]{milletari2016v}
Milletari, F., Navab, N., and Ahmadi, S.-A.
\newblock V-net: Fully convolutional neural networks for volumetric medical image segmentation.
\newblock In \emph{2016 fourth international conference on 3D vision (3DV)}, pp.\  565--571. Ieee, 2016.

\bibitem[Mnih(2013)]{MnihThesis}
Mnih, V.
\newblock \emph{Machine Learning for Aerial Image Labeling}.
\newblock PhD thesis, University of Toronto, 2013.

\bibitem[Nunez-Iglesias et~al.(2013)Nunez-Iglesias, Kennedy, Parag, Shi, and Chklovskii]{nunez2013machine}
Nunez-Iglesias, J., Kennedy, R., Parag, T., Shi, J., and Chklovskii, D.~B.
\newblock Machine learning of hierarchical clustering to segment 2d and 3d images.
\newblock \emph{PloS one}, 8\penalty0 (8):\penalty0 e71715, 2013.

\bibitem[Oktay et~al.(2018)Oktay, Schlemper, Folgoc, Lee, Heinrich, Misawa, Mori, McDonagh, Hammerla, Kainz, et~al.]{oktay2018attention}
Oktay, O., Schlemper, J., Folgoc, L.~L., Lee, M., Heinrich, M., Misawa, K., Mori, K., McDonagh, S., Hammerla, N.~Y., Kainz, B., et~al.
\newblock Attention u-net: Learning where to look for the pancreas.
\newblock \emph{arXiv preprint arXiv:1804.03999}, 2018.

\bibitem[Oner et~al.(2023)Oner, Garin, Kozi{\'n}ski, Hess, and Fua]{oner2023persistent}
Oner, D., Garin, A., Kozi{\'n}ski, M., Hess, K., and Fua, P.
\newblock Persistent homology with improved locality information for more effective delineation.
\newblock \emph{IEEE Transactions on Pattern Analysis and Machine Intelligence}, 45\penalty0 (8):\penalty0 10588--10595, 2023.

\bibitem[Portilla \& Simoncelli(2000)Portilla and Simoncelli]{portilla2000parametric}
Portilla, J. and Simoncelli, E.~P.
\newblock A parametric texture model based on joint statistics of complex wavelet coefficients.
\newblock \emph{International journal of computer vision}, 40:\penalty0 49--70, 2000.

\bibitem[Prantl et~al.(2022)Prantl, Bender, Kugelstadt, and Thuerey]{prantl2022wavelet}
Prantl, L., Bender, J., Kugelstadt, T., and Thuerey, N.
\newblock Wavelet-based loss for high-frequency interface dynamics.
\newblock \emph{arXiv preprint arXiv:2209.02316}, 2022.

\bibitem[Rahman \& Wang(2016)Rahman and Wang]{rahman2016optimizing}
Rahman, M.~A. and Wang, Y.
\newblock Optimizing intersection-over-union in deep neural networks for image segmentation.
\newblock In \emph{International symposium on visual computing}, pp.\  234--244. Springer, 2016.

\bibitem[Ronneberger et~al.(2015)Ronneberger, Fischer, and Brox]{ronneberger2015u}
Ronneberger, O., Fischer, P., and Brox, T.
\newblock U-net: Convolutional networks for biomedical image segmentation.
\newblock In \emph{Medical image computing and computer-assisted intervention--MICCAI 2015: 18th international conference, Munich, Germany, October 5-9, 2015, proceedings, part III 18}, pp.\  234--241. Springer, 2015.

\bibitem[Ross \& Doll{\'a}r(2017)Ross and Doll{\'a}r]{ross2017focal}
Ross, T.-Y. and Doll{\'a}r, G.
\newblock Focal loss for dense object detection.
\newblock In \emph{proceedings of the IEEE conference on computer vision and pattern recognition}, pp.\  2980--2988, 2017.

\bibitem[Ruan et~al.(2024)Ruan, Li, and Xiang]{ruan2024vm}
Ruan, J., Li, J., and Xiang, S.
\newblock Vm-unet: Vision mamba unet for medical image segmentation.
\newblock \emph{arXiv preprint arXiv:2402.02491}, 2024.

\bibitem[Salehi et~al.(2017)Salehi, Erdogmus, and Gholipour]{salehi2017tversky}
Salehi, S. S.~M., Erdogmus, D., and Gholipour, A.
\newblock Tversky loss function for image segmentation using 3d fully convolutional deep networks.
\newblock In \emph{International workshop on machine learning in medical imaging}, pp.\  379--387. Springer, 2017.

\bibitem[Sampat et~al.(2009)Sampat, Wang, Gupta, Bovik, and Markey]{sampat2009complex}
Sampat, M.~P., Wang, Z., Gupta, S., Bovik, A.~C., and Markey, M.~K.
\newblock Complex wavelet structural similarity: A new image similarity index.
\newblock \emph{IEEE transactions on image processing}, 18\penalty0 (11):\penalty0 2385--2401, 2009.

\bibitem[Shit et~al.(2021)Shit, Paetzold, Sekuboyina, Ezhov, Unger, Zhylka, Pluim, Bauer, and Menze]{shit2021cldice}
Shit, S., Paetzold, J.~C., Sekuboyina, A., Ezhov, I., Unger, A., Zhylka, A., Pluim, J.~P., Bauer, U., and Menze, B.~H.
\newblock cldice-a novel topology-preserving loss function for tubular structure segmentation.
\newblock In \emph{Proceedings of the IEEE/CVF conference on computer vision and pattern recognition}, pp.\  16560--16569, 2021.

\bibitem[Simoncelli et~al.(1992)Simoncelli, Freeman, Adelson, and Heeger]{simoncelli1992shiftable}
Simoncelli, E.~P., Freeman, W.~T., Adelson, E.~H., and Heeger, D.~J.
\newblock Shiftable multiscale transforms.
\newblock \emph{IEEE transactions on Information Theory}, 38\penalty0 (2):\penalty0 587--607, 1992.

\bibitem[Sirinukunwattana et~al.(2017)Sirinukunwattana, Pluim, Chen, Qi, Heng, Guo, Wang, Matuszewski, Bruni, Sanchez, et~al.]{sirinukunwattana2017gland}
Sirinukunwattana, K., Pluim, J.~P., Chen, H., Qi, X., Heng, P.-A., Guo, Y.~B., Wang, L.~Y., Matuszewski, B.~J., Bruni, E., Sanchez, U., et~al.
\newblock Gland segmentation in colon histology images: The glas challenge contest.
\newblock \emph{Medical image analysis}, 35:\penalty0 489--502, 2017.

\bibitem[Staal et~al.(2004)Staal, Abr{\`a}moff, Niemeijer, Viergever, and Van~Ginneken]{staal2004ridge}
Staal, J., Abr{\`a}moff, M.~D., Niemeijer, M., Viergever, M.~A., and Van~Ginneken, B.
\newblock Ridge-based vessel segmentation in color images of the retina.
\newblock \emph{IEEE transactions on medical imaging}, 23\penalty0 (4):\penalty0 501--509, 2004.

\bibitem[Stucki et~al.(2023)Stucki, Paetzold, Shit, Menze, and Bauer]{stucki2023topologically}
Stucki, N., Paetzold, J.~C., Shit, S., Menze, B., and Bauer, U.
\newblock Topologically faithful image segmentation via induced matching of persistence barcodes.
\newblock In \emph{International Conference on Machine Learning}, pp.\  32698--32727. PMLR, 2023.

\bibitem[Tang et~al.(2025)Tang, Liu, Zhang, Jiang, Tang, and Chen]{tang2025increase}
Tang, Q., Liu, F., Zhang, D., Jiang, J., Tang, X., and Chen, C.~P.
\newblock Increase the sensitivity of moderate examples for semantic image segmentation.
\newblock \emph{Image and Vision Computing}, 154:\penalty0 105357, 2025.

\bibitem[Vinh et~al.(2009)Vinh, Epps, and Bailey]{vinh2009information}
Vinh, N.~X., Epps, J., and Bailey, J.
\newblock Information theoretic measures for clusterings comparison: is a correction for chance necessary?
\newblock In \emph{Proceedings of the 26th annual international conference on machine learning}, pp.\  1073--1080, 2009.

\bibitem[Wang et~al.(2004)Wang, Bovik, Sheikh, and Simoncelli]{wang2004image}
Wang, Z., Bovik, A.~C., Sheikh, H.~R., and Simoncelli, E.~P.
\newblock Image quality assessment: from error visibility to structural similarity.
\newblock \emph{IEEE transactions on image processing}, 13\penalty0 (4):\penalty0 600--612, 2004.

\bibitem[Woo et~al.(2018)Woo, Park, Lee, and Kweon]{woo2018cbam}
Woo, S., Park, J., Lee, J.-Y., and Kweon, I.~S.
\newblock Cbam: Convolutional block attention module.
\newblock In \emph{Proceedings of the European conference on computer vision (ECCV)}, pp.\  3--19, 2018.

\bibitem[Yang et~al.(2020)Yang, Yang, and Tsai]{yang2020net}
Yang, H.-H., Yang, C.-H.~H., and Tsai, Y.-C.~J.
\newblock Y-net: Multi-scale feature aggregation network with wavelet structure similarity loss function for single image dehazing.
\newblock In \emph{ICASSP 2020-2020 IEEE International Conference on Acoustics, Speech and Signal Processing (ICASSP)}, pp.\  2628--2632. IEEE, 2020.

\bibitem[Yue \& Li(2024)Yue and Li]{yue2024medmamba}
Yue, Y. and Li, Z.
\newblock Medmamba: Vision mamba for medical image classification.
\newblock \emph{arXiv preprint arXiv:2403.03849}, 2024.

\bibitem[Zhao et~al.(2019)Zhao, Wang, Yang, and Cai]{zhao2019region}
Zhao, S., Wang, Y., Yang, Z., and Cai, D.
\newblock Region mutual information loss for semantic segmentation.
\newblock \emph{Advances in Neural Information Processing Systems}, 32, 2019.

\bibitem[Zhou et~al.(2018)Zhou, Rahman~Siddiquee, Tajbakhsh, and Liang]{zhou2018unet++}
Zhou, Z., Rahman~Siddiquee, M.~M., Tajbakhsh, N., and Liang, J.
\newblock Unet++: A nested u-net architecture for medical image segmentation.
\newblock In \emph{Deep Learning in Medical Image Analysis and Multimodal Learning for Clinical Decision Support: 4th International Workshop, DLMIA 2018, and 8th International Workshop, ML-CDS 2018, Held in Conjunction with MICCAI 2018, Granada, Spain, September 20, 2018, Proceedings 4}, pp.\  3--11. Springer, 2018.

\bibitem[Zhu et~al.(2021)Zhu, Wang, and Zhang]{zhu2021wavelet}
Zhu, Q., Wang, H., and Zhang, R.
\newblock Wavelet loss function for auto-encoder.
\newblock \emph{IEEE Access}, 9:\penalty0 27101--27108, 2021.

\end{thebibliography}
\bibliographystyle{icml2025}




\end{document}